\documentclass[journal]{IEEEtran}
\usepackage{subfigure}
\usepackage{multirow}
\usepackage{threeparttable}
\usepackage{enumitem}
\usepackage{bbm}
\usepackage{pifont}
\usepackage{diagbox}
\usepackage{booktabs}
\usepackage{amsmath}
\usepackage{url}
\usepackage{amssymb}
\usepackage{ragged2e}
\usepackage{pgfplots}
\usepackage{graphicx}
\usepackage{algorithm}
\usepackage{algpseudocode}

\ifCLASSOPTIONcompsoc
  \usepackage[nocompress]{cite}
\else
  \usepackage{cite}
\fi

\title{
UVTM: Universal Vehicle Trajectory Modeling with ST Feature Domain Generation
}

\author{
Yan~Lin,
Jilin~Hu,
Shengnan~Guo,
Bin~Yang,
Christian~S.~Jensen,
Youfang Lin,
Huaiyu~Wan
\thanks{Corresponding author: Huaiyu~Wan}
\IEEEcompsocitemizethanks{\IEEEcompsocthanksitem 
Yan Lin and Christian S. Jensen are with the Department of Computer Science, Aalborg University, Aalborg, 9220, Denmark.
Shengnan Guo, Youfang Lin, and Huaiyu Wan are with the Beijing Key Laboratory of Traffic Data Mining and Embodied Intelligence, School of Computer Science and Technology, Beijing Jiaotong University, Beijing 100044, China.
Jilin Hu and Bin Yang are with the School of Data Science and Engineering, East China Normal University, Shanghai 200050, China.
\protect\\
E-mail: lyan@cs.aau.dk;
jlhu@dase.ecnu.edu.cn;
guoshn@bjtu.edu.cn;
byang@dase.ecnu.edu.cn;
csj@cs.aau.dk;
yflin@bjtu.edu.cn;
hywan@bjtu.edu.cn.
}
}

\pgfplotsset{compat=1.18}
\newenvironment{customlegend}[1][]{%
        \begingroup
        \csname pgfplots@init@cleared@structures\endcsname
        \pgfplotsset{#1}%
    }{%
        \csname pgfplots@createlegend\endcsname
        \endgroup
    }%
    \def\addlegendimage{\csname pgfplots@addlegendimage\endcsname}
\definecolor{pt1min}{HTML}{1A237E}
\definecolor{pt2min}{HTML}{B71C1C}
\definecolor{pt4min}{HTML}{1B5E20}
\definecolor{wopt1min}{HTML}{01579B}
\definecolor{wopt2min}{HTML}{BF360C}
\definecolor{wopt4min}{HTML}{004D40}
\definecolor{pttte}{HTML}{E65100}
\definecolor{wopttte}{HTML}{FFD600}

\DeclareMathOperator*{\argmax}{argmax}

\newtheorem{definition}{Definition}
\newtheorem{example}{Example}

\definecolor{mutedRed}{HTML}{C62E2E}
\newcommand{\rev}[1]{#1}

\begin{document}

\maketitle

\begin{abstract}
\rev{Vehicle movement is frequently captured in the form of GPS trajectories, i.e., sequences of timestamped GPS locations. Such data is widely used for various tasks such as travel-time estimation, trajectory recovery, and trajectory prediction. A universal vehicle trajectory model could be applied to different tasks, removing the need to maintain multiple specialized models, thereby reducing computational and storage costs. However, creating such a model is challenging when the integrity of trajectory features is compromised, i.e., in scenarios where only partial features are available or the trajectories are sparse.}

\rev{
To address these challenges, we propose the Universal Vehicle Trajectory Model (UVTM), which can effectively adapt to different tasks without excessive retraining. UVTM incorporates two specialized designs. First, it divides trajectory features into three distinct domains. Each domain can be masked and generated independently to accommodate tasks with only partially available features. Second, UVTM is pre-trained by reconstructing dense, feature-complete trajectories from sparse, feature-incomplete counterparts, enabling strong performance even when the integrity of trajectory features is compromised. Experiments involving four representative trajectory-related tasks on three real-world vehicle trajectory datasets provide insight into the performance of UVTM and offer evidence that it is capable of meeting its objectives.
}
\end{abstract}


\IEEEdisplaynontitleabstractindextext
\IEEEpeerreviewmaketitle

\section{Introduction}\label{sec:introduction}
\begin{figure*}
    \centering
    \subfigure[\rev{Typical flow of trajectory modeling.}] {
        \begin{minipage}[t]{0.31\linewidth}
        \centering
        \includegraphics[width=1.0\linewidth]{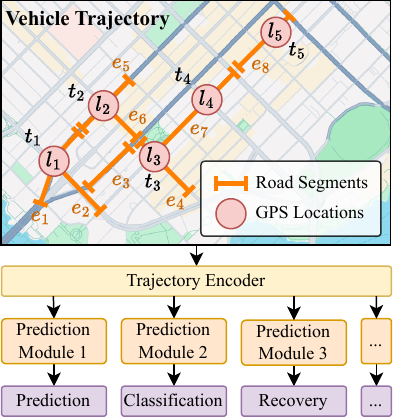}
        \end{minipage}
        \label{fig:flow}
    }
    \hfill
    \subfigure[\rev{Scenarios in which the integrity of trajectory features is compromised.}] {
        \begin{minipage}[t]{0.64\linewidth}
        \centering
        \includegraphics[width=1.0\linewidth]{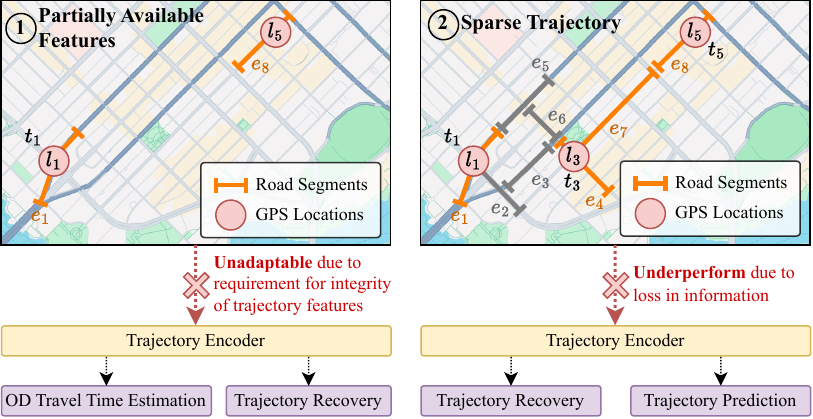}
        \end{minipage}
        \label{fig:limitations}
    }
    \caption{\rev{Overview of a typical flow of trajectory modeling, and scenarios involving incomplete or sparse trajectories that pose challenges for trajectory modeling.}}
    \label{fig:existing-methods-overview}
\end{figure*}

\IEEEPARstart {A} vehicle \rev{GPS trajectory is a sequence of timestamped GPS locations that captures the movement of a vehicle during a trip.} Figure~\ref{fig:flow} provides an example where a trajectory $\mathcal T$ captures the trip of a vehicle from $l_1$ to $l_5$. This trajectory consists of five timestamped GPS points: $\mathcal T=\langle (l_1,t_1), (l_2,t_2), \dots, (l_5,t_5) \rangle$.
Substantial spatiotemporal information can be mined from this trajectory, e.g., 1)~the travel time of the trip, $t_5-t_1$; 2)~the road segments that the trip visited, $e_1, e_2, e_4$, and $e_6$; 3)~the average travel speed on each road segment, which can reflect traffic conditions; and 4)~accelerations and decelerations that capture driving behavior.
Such information provides a rich foundation for analyzing the movement patterns of individual vehicles and traffic patterns on road networks, which in turn powers various important tasks in Intelligent Transportation Systems~(ITS). Examples of such tasks include trajectory prediction~\cite{feng2018deepmove,kong2018hst,liang2021nettraj,DBLP:journals/pvldb/FangPCDG21}, travel time estimation~\cite{yuan2020effective,fu2020compacteta,DBLP:journals/vldb/GuoYHJC20,DBLP:journals/pvldb/PedersenYJ20}, anomaly detection~\cite{liu2020online,han2022deeptea}, and trajectory recovery~\cite{DBLP:conf/aaai/XiaQ0XSGL21,DBLP:conf/kdd/RenRL0ML021,DBLP:journals/corr/abs-2211-13234}.

\rev{
Given the diverse applicability of trajectories, the same trajectory dataset is often utilized for various tasks simultaneously. For example, a ride-hailing company may use its collected taxi trajectory dataset to: 1) estimate travel times to improve customer ETAs and driver dispatching; 2) recover missing GPS points to maintain data quality despite signal loss; and 3) predict future trajectories to pre-position drivers in high-demand areas.
Traditionally, individual models are created and trained for each specific task. Yet, training and storing multiple models for different tasks adversely affect computational and storage efficiency --- a large city generating millions of trajectories daily would require maintaining and updating several large models simultaneously.
}
To address this, there is a pressing need to develop a universal vehicle trajectory model that can be trained once on a dataset and effectively address various types of tasks.

\rev{
Existing efforts~\cite{DBLP:conf/icde/LiZCJW18,fu2020trembr,jiang2022self,DBLP:conf/kdd/YangHGYJ23,yu2024bigcity,zhou2024plm4traj,wei2025path} have explored universal trajectory models using either representation learning or multi-task learning. In representation learning, a trajectory encoder is pre-trained to map each trajectory into an embedding vector, which is then used for each specific task via an attached prediction module. In multi-task learning, a trajectory encoder and multiple task-specific prediction modules are trained together under multi-task supervision. Figure~\ref{fig:flow} outlines the workflow for both categories.
However, these existing methods do not adequately address two scenarios where the integrity of trajectory features is compromised, creating challenges for adapting universal trajectory models. We elaborate on these scenarios below.
}

\rev{\textbf{Scenario 1: Tasks in which trajectory features are only partially available.}}
\rev{
There is a wide range of trajectory-related tasks that involve varying arrangements of input and output features. Some tasks involve only partially available trajectory features as input. For instance, in origin-destination (OD) travel time estimation~\cite{DBLP:journals/pacmmod/LinWHGYLJ23}, only the origin, destination, and departure time of a trajectory are available prior to its occurrence. Similarly, in trajectory recovery~\cite{DBLP:conf/gis/ElshrifIM22}, there are often long spans of missing trajectory points among the known ones. When performing these tasks, universal trajectory models that rely on the integrity of the trajectory features are unable to adapt, as illustrated in Figure~\ref{fig:limitations}, scenario~\ding{172}.
}

\rev{
Existing methods typically have strict requirements for the availability of trajectory features, limiting their adaptability in the above scenario.
For example, t2vec~\cite{DBLP:conf/icde/LiZCJW18} requires the coordinate and timestamp of each trajectory point, while START~\cite{jiang2022self} and BIGCity~\cite{yu2024bigcity} require the road segment and timestamp for each point. Consequently, these methods are unable to adapt to tasks such as OD travel time estimation and trajectory recovery, where their requirements regarding feature availability cannot be fulfilled.
}

\rev{\textbf{Scenario 2: Sparse trajectories with large sampling intervals.}}
\rev{
Sparse trajectories are common in real-world trajectory datasets for two reasons. First, due to factors such as signal loss and device malfunctions, trajectories collected from vehicles can have large sampling intervals. Second, to reduce storage requirements and improve computational efficiency, trajectories can be re-sampled with larger intervals after collection. Yet, sparse trajectories lose information compared to their dense counterparts, which negatively impacts the performance of universal trajectory models if not properly addressed.
}

\rev{
Figure~\ref{fig:limitations}, scenario~\ding{173} gives an example. A sparse trajectory $\mathcal T'=\{(l_1,t_1),(l_3,t_3),(l_5,t_5)\}$ is re-sampled from the dense trajectory $\mathcal T$. When a model processes $\mathcal T'$, information not explicitly captured in $\mathcal T'$ can be overlooked. Specifically, distinguishing whether the vehicle traverses road segments $e_5, e_6$ or $e_2, e_3$ between points $(l_1,t_1)$ and $(l_3,t_3)$ is challenging given only $\mathcal T'$.
}

To effectively address the above scenarios, we propose the \textit{\underline{U}niversal \underline{V}ehicle \underline{T}rajectory \underline{M}odel} (\textbf{UVTM}). UVTM is designed to be trained once and then adjusted to effectively address various trajectory-related tasks, facilitating its versatility in ITS applications. 
\rev{UVTM incorporates two specialized designs that excel at handling scenarios in which the integrity of trajectory features is compromised. First, we divide the spatiotemporal features in trajectories into three distinct domains: spatial, temporal, and road. Each domain can be masked and generated independently to handle scenarios in which trajectory features are only partially available, and meet the output requirements of the task at hand. Second, we pre-train UVTM to enable it to extract spatiotemporal and road segment features embedded in the dense, feature-complete trajectories, even when presented with incomplete or sparse trajectories. This pre-training ensures that UVTM consistently delivers strong performance in scenarios in which the integrity of trajectory features is compromised.}
In summary, the primary contributions of the paper are as follows.
\begin{itemize}[leftmargin=*]
    \item \rev{We propose the UVTM, a versatile universal vehicle trajectory model. UVTM can be trained once and effectively adapt to multiple trajectory-related tasks without excessive retraining or additional modules, and effectively handles scenarios in which the integrity of trajectory features is compromised.}
    \item \rev{We divide the features in trajectories into three domains, allowing each domain to be masked and generated separately based on the specific availability of trajectory features and the output requirements of each task. This flexibility enables the UVTM to adapt to various types of tasks, particularly those involving incomplete or sparse trajectories.}
    \item \rev{We pre-train UVTM by reconstructing dense, feature-complete trajectories using their sparse, feature-incomplete counterparts. This process enhances UVTM's robustness to sparsity and incompleteness of trajectory features.}
    \item \rev{We conduct extensive experiments on three real-world vehicle trajectory datasets and four representative tasks, offering insight into the performance properties of the proposed model, and providing evidence that the model is capable of meeting its design goals.}
\end{itemize}

\section{Related Work} \label{sec:related-work}

\subsection{Task-specific Trajectory Models}
Task-specific models are designed and trained for particular tasks, offering simpler implementations for many trajectory-related applications.

For travel time estimation, path-based approaches including WDR~\cite{wang2018learning}, DeepTTE~\cite{wang2018will}, DeepETA~\cite{wu2019deepeta}, WDDRA~\cite{gan2021travel}, and DRTTE~\cite{yang2022multitask} predict travel time using the entire travel path. Conversely, origin-destination-based approaches including TEMP~\cite{wang2019simple}, ST-NN~\cite{jindal2017unified}, MURAT~\cite{li2018multi}, DeepOD~\cite{yuan2020effective}, and DOT~\cite{DBLP:journals/corr/abs-2307-03048} estimate travel time by considering only the origin, destination, and departure time.
For trajectory recovery, TrImpute~\cite{DBLP:conf/gis/ElshrifIM22} is a non-learning-based method. Meanwhile, AttnMove~\cite{DBLP:conf/aaai/XiaQ0XSGL21}, MTrajRec~\cite{DBLP:conf/kdd/RenRL0ML021}, and RNTrajRec~\cite{DBLP:journals/corr/abs-2211-13234} are learning-based methods that use a seq2seq framework~\cite{DBLP:conf/nips/SutskeverVL14}.
For trajectory prediction, models such as DeepMove~\cite{feng2018deepmove}, HST-LSTM~\cite{kong2018hst}, and ACN~\cite{miao2020predicting} rely on recurrent neural networks~\cite{hochreiter1997long} to capture sequential patterns. \rev{PreCLN~\cite{yan2022precln} further integrates contrastive learning~\cite{DBLP:conf/icml/ChenK0H20} to improve prediction accuracy.}

\rev{
Although task-specific trajectory models can perform well for their respective tasks, they cannot be easily adapted to tasks for which they were not originally designed. This limitation often leads to the need for separate models for different tasks, reducing both computational and storage efficiency.
}

\subsection{Universal Trajectory Models}
To address the limitations of task-specific models, growing attention is being paid to universal trajectory models capable of handling multiple tasks.

Among these, trajectory2vec~\cite{DBLP:conf/ijcnn/YaoZZHB17} constructs behavior sequences from trajectories to extract key information, then compresses each sequence into an embedding with an auto-encoding framework~\cite{hinton2006reducing}. t2vec~\cite{DBLP:conf/icde/LiZCJW18} adopts a denoising auto-encoder to improve robustness against noise. Trembr~\cite{fu2020trembr} employs auto-encoders to capture and integrate road network and temporal information. SML~\cite{DBLP:journals/kbs/ZhouDGWZ21} introduces a contrastive predictive coding framework~\cite{DBLP:journals/corr/abs-1807-03748} for trajectory embedding. START~\cite{jiang2022self} combines masked language modeling~\cite{DBLP:conf/naacl/DevlinCLT19} with SimCLR~\cite{DBLP:conf/icml/ChenK0H20} to enrich its representation capabilities. \rev{PathLLM~\cite{wei2025path} leverages large language models to learn multi-modal path representations, which can also be applied to trajectory learning. PLM4Traj~\cite{zhou2024plm4traj} and BIGCity~\cite{yu2024bigcity} further use pre-trained language models in a multi-task framework, enabling different trajectory-related tasks to be performed via task-specific prompts.}

\rev{Although these universal models offer flexibility, they are still limited when handling tasks involving incomplete trajectory features, and often underperform with sparse trajectories.}

\section{Preliminaries} \label{sec:preliminaries}

\subsection{Definitions}
\begin{definition}
[Road Network]
A road network is modeled as a directed graph $\mathcal G=(\mathcal V, \mathcal E)$, where $\mathcal V$ is a set of nodes, each node $v_i\in \mathcal V$ models an intersection between road segments or the end of a segment, and $\mathcal E$ is a set of edges, each edge $e_i\in\mathcal E$ models a road segment linking two nodes.
An edge is given by starting and ending nodes: $e_i=(v_j,v_k)$.
\end{definition}

\begin{definition}
[GPS Trajectory]
A GPS trajectory $\mathcal T$ is a sequence of timestamped point locations: $\mathcal T=\langle (l_1,t_1), (l_2,t_2), \dots, (l_{|\mathcal T|},t_{|\mathcal T|}) \rangle$, where $l_i=(l^\mathrm{lng}_i,l^\mathrm{lat}_i)$ are the spatial coordinates of the $i$-th location, $l^\mathrm{lng}_i$ and $l^\mathrm{lat}_i$ denote longitude and latitude, respectively, and timestamp $t_i$ is the time at which $l_i$ is visited.
\end{definition}

\rev{
In this work, we focus on vehicle GPS trajectories. Each trajectory records a vehicle's movement from an origin to a destination. Trajectories are also used in other applications (e.g., people visiting points of interest~\cite{wang2023would,DBLP:conf/gis/0001VS22,DBLP:conf/nips/Gong0ZLH00LW24,zhang2024large} or vehicle behavior at intersections~\cite{DBLP:conf/cvpr/RoweED023,DBLP:conf/iccv/ChenWSLHGCH23,DBLP:conf/iros/LiZXTLDTZ24}), but these are beyond this paper's scope.
}

\begin{definition}
[Sampling Interval]
For a trajectory $\mathcal{T}$, its sampling interval $\eta$ is the time gap between consecutive points, i.e., $\eta = t_i - t_{i-1}$ for $i \in \{2,3,\dots,|\mathcal T|\}$. Given a dense trajectory with a small $\eta$, one can obtain a sparse trajectory by re-sampling with a larger interval $\mu$.
\end{definition}

We assume consistent sampling intervals in our experiments for ease of evaluation, although the proposed model can be applied to trajectories with varying intervals.

\begin{definition}
[Map-matched Trajectory] \label{def:map-matched-trajectory}
A map-matching algorithm~\cite{chao2020survey} projects a trajectory $\mathcal{T}$ onto the underlying road network $\mathcal{G}$. The resulting map-matched trajectory can be denoted as $\widetilde{\mathcal T}=\langle (\widetilde{l}_1,t_1), (\widetilde{l}_2,t_2), \dots, (\widetilde{l}_{|\mathcal T|},t_{|\mathcal T|}) \rangle$, where $\widetilde{l}_i = (e_i, r_i)$, $e_i \in \mathcal{E}$ is the road segment that corresponds to $l_i$, and $r_i$ is the fraction of $e_i$ that has been traveled.
\end{definition}

\begin{figure}
    \centering
    \includegraphics[width=1.0\linewidth]{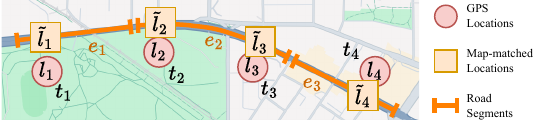}
    \caption{\rev{A vehicle trajectory and its map-matched counterpart.}}
    \label{fig:map-matched-trajectory}
\end{figure}

\begin{example}
Figure~\ref{fig:map-matched-trajectory} illustrates a trajectory $\mathcal T=\langle (l_1,t_1), (l_2,t_2), (l_3,t_3), (l_{4},t_{4}) \rangle$, where the GPS locations are denoted in circles. Map-matching produces $\widetilde{\mathcal T}=\langle (\tilde{l}_1,t_1), (\tilde{l}_2,t_2), (\tilde{l}_3,t_3), (\tilde{l}_{4},t_{4}) \rangle$, where the map-matched locations are in squares. For example, $l_2$ is matched to a point $\tilde{l}_2$ located midway along the road segment $e_2$. We denote by $r_2$ the fraction of $e_2$ traveled by time $t_2$, so $\widetilde{l}_2 = (e_2, r_2)$.
\end{example}

\subsection{Problem Statement}
\noindent \textbf{Universal Vehicle Trajectory Modeling.}
The objective is to construct a universal vehicle trajectory model $f_\theta$, where $\theta$ denotes a set of learnable parameters.
During evaluation, $f_\theta$ takes a certain arrangement of trajectory $\mathcal T$ as input and generates an output tailored to the particular task at hand, denoted as $\hat Y = f_\theta(\mathrm{arrange}(\mathcal T))$.
For example, for OD travel time estimation, $\mathrm{arrange}(\mathcal T)$ extracts the origin, destination, and departure time of $\mathcal T$, and $\hat Y$ is the estimated travel time; for trajectory prediction, $\mathrm{arrange}(\mathcal T)$ retains the historical part of $\mathcal T$, and $\hat Y$ is the predicted future part of $\mathcal T$.

\begin{figure*}[t]
    \centering
    \subfigure[\rev{Representing a trajectory point as a tuple.}] {
        \begin{minipage}[t]{0.17\linewidth}
        \centering
        \includegraphics[width=1.0\linewidth]{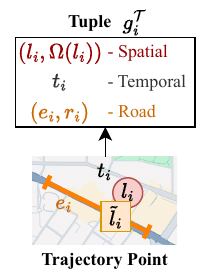}
        \end{minipage}
        \label{fig:feature-domains}
    }
    \hspace{0.02\linewidth}
    \subfigure[\rev{Auto-regressive generation of blocks of tuples for input tuples with masked domains. Each input tuple corresponds to a generated block of tuples.}] {
        \begin{minipage}[t]{0.7\linewidth}
        \centering
        \includegraphics[width=1.0\linewidth]{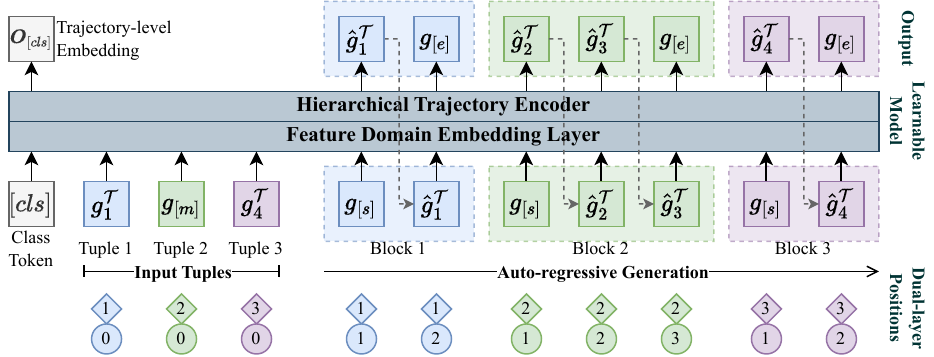}
        \end{minipage}
        \label{fig:autoregressive-generation-of-tuples}
    }
    \caption{\rev{Two core components of the proposed UVTM: tuple of feature domains and auto-regressive generation of the tuples.}}
    \label{fig:overall-framework}
\end{figure*}

\section{Methodology} \label{sec:methodology}

\subsection{Overall Framework}
\label{sec:overall-framework}

The overall framework of the proposed UVTM is illustrated in Figure~\ref{fig:overall-framework}. \rev{We first divide trajectory features into three domains: spatial, temporal, and road, as shown in Figure~\ref{fig:feature-domains}. Each point in the trajectory is represented as a \textit{tuple} containing these three domains. We denote the tuple for the $i$-th point in trajectory $\mathcal T$ as $g_i^{\mathcal T}$.}

\rev{
Each feature domain in $g_i^{\mathcal T}$ can be independently masked and generated. Specifically, any domain may be replaced with the mask token $[m]$ to indicate it requires generation. Likewise, the tuple $g_{[m]}$ (with all domains filled with the mask token $[m]$) represents an entire sub-trajectory that needs to be generated. For each input tuple containing $[m]$, a sequence of tuples is auto-regressively generated, forming a \textit{block} of generated tuples. The process begins by appending the tuple $g_{[s]}$ (filled with the start token $[s]$) to the input, then iteratively appending each newly generated tuple back into the input to produce the next one. The process concludes when the tuple $g_{[e]}$ (filled with the end token $[e]$) is produced. This design allows UVTM to handle diverse tasks by accepting many task-specific input configurations and producing the required output.
Figure~\ref{fig:autoregressive-generation-of-tuples} illustrates an example of three blocks generated for three input tuples: $g_1^\mathcal T$ and $g_4^\mathcal T$ each have one masked domain (generated blocks 1 and 3, respectively), while $g_{[m]}$ represents a missing sub-trajectory (generated block 2) that bridges $g_1^\mathcal T$ and $g_4^\mathcal T$.
}

\rev{
To enable the model to extract and correlate information across these feature domains, we introduce the feature domain embedding layer and the hierarchical trajectory encoder. Furthermore, to improve the model’s robustness when dealing with incomplete trajectories, we pre-train it to reconstruct dense, feature-complete trajectories from sparse, feature-incomplete ones. The following sections detail UVTM’s design and modules.
}

\subsection{Feature Domains}
\rev{
To address the varied availability of trajectory features in different tasks, we divide these features into three domains. Each domain can be masked and generated separately, depending on which features are available and what outputs a task requires.
}

\subsubsection{Tuple of Feature Domains}
Given the $i$-th point $(l_i,t_i)$ of trajectory $\mathcal T$, we split its spatiotemporal features into three domains: spatial, temporal, and road.

\rev{
\textbf{Spatial domain} contains the coordinate $l_i$, as well as the set $\Omega(l_i)$ of road segments lying within $\delta$ meters of $l_i$. Including $\Omega(l_i)$ improves the model’s understanding of how coordinates relate to nearby road segments.
}

\rev{
\textbf{Temporal domain} includes the timestamp $t_i$ in seconds. We normalize $t_i$ by subtracting $t_0$ from it.
}

\rev{
\textbf{Road domain} consists of the road segment $e_i$ and fraction $r_i$ from the map-matched point $\tilde l_i$, obtained by map-matching $l_i$ according to Definition~\ref{def:map-matched-trajectory}.
}

We then convert $(l_i,t_i)$ into a \textit{tuple} $g_i^{\mathcal T}$ containing these three domains:
\begin{equation}
    g_i^{\mathcal T} = ((l_i,\Omega(l_i)), t_i, (e_i,r_i))
    \label{eq:tuple-of-feature-domains}
\end{equation}

\subsubsection{Auto-regressive Generation}
\label{sec:generation-of-feature-domains}
\rev{
To generate a feature domain in $g_i^{\mathcal T}$, we replace that domain with the mask token $[m]$ and provide the tuple to the model as input. The model then auto-regressively generates a series of tuples, forming a \textit{block}. The generation begins by adding a tuple $g_{[s]}=([s],[s],[s])$ to the input. The model outputs the generated tuple $\hat g_i^{\mathcal T}$, which is then appended to the input, and the process continues until the tuple $g_{[e]}=([e],[e],[e])$ is produced. An example of this is shown by the input tuple $g_1^\mathcal T$ and the block $\langle \hat g_1^\mathcal T, g_{[e]} \rangle$ in Figure~\ref{fig:autoregressive-generation-of-tuples}.
}

\rev{
If all feature domains of a consecutive sub-trajectory need to be generated, we use a single tuple $g_{[m]}=([m],[m],[m])$. The generation follows the same auto-regressive process, except the model repeatedly appends each newly generated tuple to the input until it decides to stop by producing $g_{[e]}$. As illustrated by $g_{[m]}$ and its generated block $\langle \hat g_2^\mathcal T, \hat g_3^\mathcal T, g_{[e]} \rangle$ in Figure~\ref{fig:autoregressive-generation-of-tuples}, the process can produce sub-trajectories of varying lengths, which is useful for tasks such as trajectory recovery and prediction.
}

\rev{
When multiple masked tuples are provided, they are ordered into an input sequence. The model then generates their corresponding blocks in turn. Figure~\ref{fig:autoregressive-generation-of-tuples} shows an example in which three blocks are generated for three input tuples.
}

\subsubsection{Dual-layer Positions}
Since each input tuple corresponds to a generated block, we follow the practice of GLM~\cite{DBLP:conf/acl/DuQLDQY022} and employ dual-layer positions.

\rev{
\textbf{First layer} ($\mathcal P_i^1$) marks the relationship between each input tuple $g_i^{\mathcal T}$ and its generated block. Specifically, $g_i^{\mathcal T}$ and its generated block share the same first-layer position $i$.
}

\rev{
\textbf{Second layer} ($\mathcal P_i^2$) marks the order of tuples within each block. For input tuples, this position is always $0$. For a block of $N$ generated tuples, the positions are assigned from $1$ to $N$ in order.
}

\subsubsection{\rev{Class Token}}
\rev{
To handle trajectory-level tasks such as similar search, we introduce a class token $[cls]$ at the start of the input sequence. The model’s output state corresponding to $[cls]$ is treated as the embedding of the trajectory, detailed in later sections.
}

\subsubsection{Feature Domain Embedding Layer}
\label{sec:spatiotemporal-feature-encoding}

\begin{figure}[t]
    \centering
    \subfigure[\rev{Embedding of feature domains with complete features.}] {
        \begin{minipage}[t]{0.425\linewidth}
        \centering
        \includegraphics[width=1.0\linewidth]{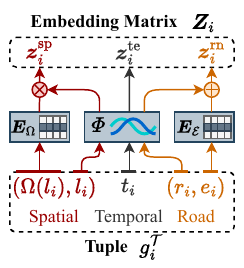}
        \end{minipage}
        \label{fig:embed-num}
    }
    \hfill
    \subfigure[\rev{Embedding of special tokens.}] {
        \begin{minipage}[t]{0.505\linewidth}
        \centering
        \includegraphics[width=1.0\linewidth]{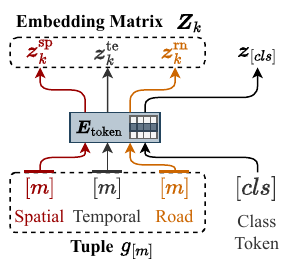}
        \end{minipage}
        \label{fig:embed-token}
    }
    \caption{\rev{Pipeline of the feature domain embedding layer. Continuous features are encoded with learnable Fourier features, while discrete features use index-fetching embedding modules.}}
    \label{fig:feature-encoding}
\end{figure}

To enable the model to extract information from the three feature domains, we propose an embedding layer that projects domains into the latent space, as illustrated in Figure~\ref{fig:feature-encoding}.

\rev{
\textbf{Continuous features.} Such features in the tuple $g_i^{\mathcal T}$, namely $l_i^\mathrm{lng}$, $l_i^\mathrm{lat}$, $t_i$, and $r_i$, have periodic characteristics. We encode these features using learnable Fourier features~\cite{tancik2020fourier,li2021learnable}, which rely on trigonometric functions that preserve periodic characteristics. Given a continuous feature $x\in \mathbb R$, the encoding module $\varPhi$ projects it into a $d$-dimensional latent space:}
\begin{equation}
    \varPhi(x) = \boldsymbol{W}_\varPhi[\cos(x \boldsymbol{v}_\varPhi) \Vert \sin(x \boldsymbol{v}_\varPhi)], x \in \{ l_i^\mathrm{lng}, l_i^\mathrm{lat}, t_i, r_i \},
\label{eq:continuous-numerical-feature-encoding}
\end{equation}
\rev{
where $\boldsymbol{v}_\varPhi \in \mathbb R^{d/2}$ and $\boldsymbol{W}_\varPhi \in \mathbb R^{d\times d}$ are learnable parameters. Each of the four continuous features has its own set of parameters.
}

\rev{
\textbf{Discrete features.} The road segment $e_i$, the segments in $\Omega(l_i)$, and the special tokens $\{[m],[s],[e],[cls]\}$ are discrete. For these, we use index-fetching embedding modules with learnable matrices $\boldsymbol E_\mathcal E \in \mathbb R^{|\mathcal E| \times d}$, $\boldsymbol E_\Omega \in \mathbb R^{|\mathcal E| \times d}$, and $\boldsymbol E_\text{token} \in \mathbb R^{4 \times d}$. For instance, the embedding for $e_i$ is the $e_i$-th row of $\boldsymbol E_\mathcal E$, denoted $\boldsymbol E_\mathcal E(e_i)$.
}

\rev{
We gather these embedding vectors to form an embedding matrix for each tuple $g_i^{\mathcal T}$. Specifically,
}
\begin{equation}
\begin{split}
    {\boldsymbol z'}^\mathrm{sp}_i &= \varPhi(l_i^\mathrm{lng}) + \varPhi(l_i^\mathrm{lat}) \\ 
    \boldsymbol z^\mathrm{sp}_i &= {\boldsymbol z'}^\mathrm{sp}_i + \mathrm{MultiHead}({\boldsymbol z'}^\mathrm{sp}_i, \boldsymbol E_\Omega(l_i), \boldsymbol E_\Omega(l_i)) \\
    \boldsymbol z^\mathrm{te}_i &= \varPhi(t_i) \\ 
    \boldsymbol z^\mathrm{rn}_i &= \boldsymbol E_\mathcal E(e_i) + \varPhi(r_i),
\end{split}
\label{eq:three-aspects-latent-vectors}
\end{equation}
\rev{
where $\mathrm{MultiHead}$ is the multi-head attention from the Transformer~\cite{DBLP:conf/nips/VaswaniSPUJGKP17}, and $\boldsymbol E_\Omega(l_i)$ is the set of embedding vectors for the segments in $\Omega(l_i)$. If any feature domain is a special token (e.g., $[m]$), we simply use its token embedding.
}
Finally, the embedding matrix for the tuple $g_i^{\mathcal T}$ is $\boldsymbol Z_i = \langle \boldsymbol z^\mathrm{sp}_i, \boldsymbol z^\mathrm{te}_i, \boldsymbol z^\mathrm{rn}_i \rangle \in \mathbb R^{3\times d}$.

\begin{figure}[t]
    \centering
    \includegraphics[width=1.0\linewidth]{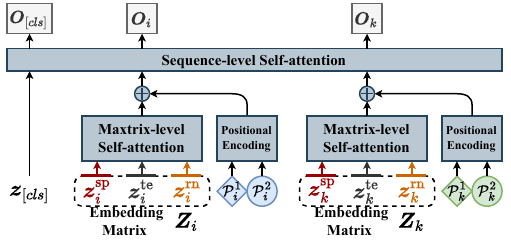}
    \caption{\rev{Architecture of the hierarchical attention in the trajectory encoder.}}
    \label{fig:trajectory-encoder}
\end{figure}

\subsection{Hierarchical Trajectory Encoder} \label{sec:trajectory-encoder}
\rev{
We now describe how UVTM models correlations among tuples and carries out the generation process in Section~\ref{sec:generation-of-feature-domains}. The trajectory encoder takes a sequence of matrices $\boldsymbol Z_i$ and outputs a sequence of tuples $\hat g_i^{\mathcal T}$.
}

\subsubsection{Hierarchical Attention}
\rev{
We first capture correlations among the three domains inside each tuple by applying a matrix-level self-attention on $\boldsymbol Z_i$, where $\boldsymbol Z_i$ is treated as a length-3 sequence. We then use mean pooling and positional encoding to obtain the hidden state $\boldsymbol h_i$:
}
\begin{equation}
    \boldsymbol h_i = \mathrm{Mean}(\mathrm{MultiHead}(\boldsymbol Z_i, \boldsymbol Z_i, \boldsymbol Z_i)) + 
    \mathrm{PE}(\mathcal P^{1}_i) + \mathrm{PE}(\mathcal P^{2}_i),
\label{eq:tuple-level-self-attention}
\end{equation}
where $\mathrm{PE}$ is the positional encoding from the Transformer.

\rev{
During the generation process in Section~\ref{sec:generation-of-feature-domains}, we compute $\boldsymbol h_i$ for each input tuple and prepend the class token embedding $\boldsymbol z_{[cls]}$ to form the sequence $\boldsymbol H=\langle \boldsymbol z_{[cls]}, \boldsymbol h_1, \boldsymbol h_2, \ldots\rangle$. We then apply sequence-level self-attention:
}
\begin{equation}
\begin{split}
    \boldsymbol H' &= \mathrm{MultiHead}(\boldsymbol H, \boldsymbol H, \boldsymbol H) \\
    \boldsymbol O' &= \mathrm{LayerNorm}(\boldsymbol H' + \boldsymbol H) \\
    \boldsymbol O &= \mathrm{LayerNorm}(\mathrm{FFN}(\boldsymbol O') + \boldsymbol O'),
\end{split}
\label{eq:sequence-level-self-attention}
\end{equation}
\rev{
where $\mathrm{FFN}$ is a two-layer feed-forward network, and $\boldsymbol O$ is the sequence of output states.
A causal mask is applied to the $\mathrm{MultiHead}$ for the portion of $\boldsymbol H$ that corresponds to generated blocks to prevent information leakage.
}


\subsubsection{Output Layer}
\rev{
The output state $\boldsymbol O_{[cls]} \in \mathbb R^d$ for the class token $[cls]$ directly serves as the trajectory embedding. For the output states that correspond to generated tuples, we use four fully connected networks to obtain each feature domain. Formally, for the $i$-th output state $\boldsymbol O_i$:
}
\begin{equation}
\begin{split}
    \hat l_i &= \boldsymbol W_c \boldsymbol O_i + \boldsymbol b_c \\
    \hat t_i &= \boldsymbol W_t \boldsymbol O_i + \boldsymbol b_t \\
    \hat e_i &= \argmax(p(\hat{e_i})),~p(\hat{e_i}) = \mathrm{Softmax}(\boldsymbol W_e \boldsymbol O_i + \boldsymbol b_e) \\
    \hat r_i &= \boldsymbol W_r \boldsymbol O_i + \boldsymbol b_r,
\end{split}
\end{equation}
where $\boldsymbol W_c \in \mathbb R^{2\times d}$, $\boldsymbol W_t \in \mathbb R^{1\times d}$, $\boldsymbol W_e \in \mathbb R^{(|\mathcal E| + 1)\times d}$, and $\boldsymbol W_r \in \mathbb R^{1\times d}$ along with $\boldsymbol b_c \in \mathbb R^2$, $\boldsymbol b_t \in \mathbb R$, $\boldsymbol b_e \in \mathbb R^{(|\mathcal E| + 1)}$, and $\boldsymbol b_r \in \mathbb R$ are learnable parameters.
The generated tuple is:
\begin{equation}
\hat g_i^{\mathcal T}=((\hat l_i, \Omega(\hat l_i)), \hat t_i, (\hat e_i, \hat r_i))
\end{equation}
If $\hat{e_i} = [e]$, we set $\hat g_i^{\mathcal T}=g_{[e]}$.

\rev{
Given the ground-truth tuple $g_i^{\mathcal T}$, the loss function for supervision is:
}
\begin{equation}
\resizebox{0.85\linewidth}{!}{\begin{math}
    \mathcal L_i = \mathbbm 1^{[e]}_i (\frac{1}{2}\Vert \hat{l_i} - l_i \Vert_2 + \Vert \hat{t_i} - t_i \Vert_2 + \Vert \hat{r_i} - r_i \Vert_2)
    - \log p(\hat{e_i}),
\end{math}}
\label{eq:tuple-loss-function}
\end{equation}
\rev{
where $\mathbbm 1^{[e]}_i$ is $0$ if $g_i^{\mathcal T}=g_{[e]}$ and $1$ otherwise.
}

\subsection{Pre-training}
\label{sec:pre-training}

\rev{
To improve the model’s performance when handling sparse or incomplete trajectories, we pre-train it by 1) reconstructing dense, feature-complete trajectories from their sparse, feature-incomplete counterparts, and 2) maximizing the similarity between the embeddings of the same trajectory's sparse and dense counterparts and minimizing the similarity between the embeddings of different trajectories.
}

\subsubsection{Reconstruction Procedure}
\label{sec:reconstruction-procedure}
We start from a dense trajectory $\mathcal T$ where the sampling interval $\eta$ does not exceed 15 seconds. We apply the Fast Map Matching (FMM) algorithm~\cite{yang2018fast} to obtain its map-matched counterpart $\widetilde{\mathcal T}$.

\rev{
To emulate sparse trajectories with longer sampling intervals, we resample $\mathcal T$ at a larger interval $\mu$ (with $\mu > \eta$ and $\mu$ divisible by $\eta$) to obtain its sparse counterpart $\mathcal T^\mu$. Formally:
}
\begin{equation}
\resizebox{0.9\linewidth}{!}{\begin{math}
    \mathcal T^\mu = \langle (l_1, t_1), (l_{1+\mu/\eta}, t_{1+\mu/\eta}), (l_{1+2\mu/\eta}, t_{1+2\mu/\eta}), \dots, (l_{|\mathcal T|}, t_{|\mathcal T|}) \rangle
\end{math}}
\label{eq:resampled-trajectory}
\end{equation}
\rev{
We randomly select $\mu$ from 1\,min, 2\,min, and 4\,min. This strategy helps the model handle sparse trajectories with different sampling intervals.
To further simulate incomplete features, we randomly decide (with $\varphi\%$ probability) whether to remove either the coordinate $l_i$ or the time $t_i$ of each point in $\mathcal T^\mu$.
}

\rev{
Next, we compare each point $(l_i, t_i)$ in $\mathcal T^\mu$ with its corresponding point $(\tilde l_i, t_i)$ in $\widetilde{\mathcal T}$. Since the road domain is absent, we represent this missing domain by $[m]$. Thus, the tuple can be written as $g_i^{\mathcal T^\mu} = ((l_i, \Omega(l_i)), t_i, [m])$.
If the point has its coordinate removed, we write it as $g_{i}^{\mathcal T^\mu}=([m], t_i, [m])$. If its time is removed, we write $g_{i}^{\mathcal T^\mu}=((l_i, \Omega(l_i)), [m], [m])$.
Meanwhile, the sub-trajectory of $\mathcal T$ lying between consecutive points in $\mathcal T^\mu$ is also missing and is represented by a single tuple $g_{[m]}$.
}

\rev{
Our goal is to reconstruct the missing feature domains of $\mathcal T^\mu$, effectively generating all feature domains in the original $\mathcal T$ and the map-matched counterpart $\widetilde{\mathcal T}$. To achieve this, we provide the model with an input sequence of the form:
}
\begin{equation}
\text{input}=\langle g_1^{\mathcal T^\mu},g_{[m]},g_{1+\mu/\eta}^{\mathcal T^\mu},g_{[m]},g_{1+2\mu/\eta}^{\mathcal T^\mu},\dots,g_{|\mathcal T^\mu|}^{\mathcal T^\mu} \rangle
\end{equation}
\rev{
The model then generates a block of tuples for each input tuple as described in Section~\ref{sec:generation-of-feature-domains}. Specifically, for each $g_i^{\mathcal T^\mu}$, the model generates the block $\langle \hat g_i^{\mathcal T^\mu}, g_{[e]} \rangle$, and for each $g_{[m]}$, it generates the block representing the sub-trajectory in $\mathcal T$ that falls between the relevant points in $\mathcal T^\mu$, ending in $g_{[e]}$.
}

\rev{
To allow the model to extract bidirectional correlations from $\mathcal T$, we follow the block-shuffling approach in GLM~\cite{DBLP:conf/nips/YangDYCSL19}. That is, we shuffle the generation order of blocks during pre-training. For example, an output sequence:
}
\begin{equation}
\text{output}=\langle \hat g_j^\mathcal T, \hat g_{j+1}^\mathcal T, g_{[e]},  \hat g_i^\mathcal T, \hat g_{i+1}^\mathcal T, g_{[e]} \rangle
\end{equation}
\rev{
is the result of the block $\langle \hat g_j^\mathcal T, \hat g_{j+1}^\mathcal T, g_{[e]} \rangle$ being generated first, followed by $\langle \hat g_i^\mathcal T, \hat g_{i+1}^\mathcal T, g_{[e]} \rangle$.
}

\subsubsection{\rev{Contrastive Learning}}
\rev{
To train the trajectory embedding $\boldsymbol O_{[cls]}$, we adopt a contrastive loss. Given a batch of dense trajectories $\mathbb T$, each dense trajectory $\mathcal T$ pairs with its sparse counterpart $\mathcal T^\mu$ (derived from Section~\ref{sec:reconstruction-procedure}). Denote the embedding of $\mathcal T$ by $\boldsymbol O_{[cls]}^{\mathcal T}$ and that of $\mathcal T^\mu$ by $\boldsymbol O_{[cls]}^{\mathcal T^\mu}$. We treat these as a positive pair and treat the sparse counterparts of all other trajectories in $\mathbb T$ as negative samples. We define the InfoNCE loss~\cite{DBLP:journals/corr/abs-1807-03748} as follows:
}
\begin{equation}
\label{eq:cl-loss}
\rev{
\mathcal L_{\mathcal T}^\mathrm{CL} 
= -\log
\frac{
\exp\Bigl(\mathrm{sim}\bigl(\boldsymbol O_{[cls]}^{\mathcal T}, \boldsymbol O_{[cls]}^{\mathcal T^\mu}\bigr)/\tau\Bigr)
}{
\sum_{\mathcal T' \in \mathbb T} 
\exp\Bigl(\mathrm{sim}\bigl(\boldsymbol O_{[cls]}^{\mathcal T}, \boldsymbol O_{[cls]}^{\mathcal T'^\mu}\bigr)/\tau\Bigr)
},
}
\end{equation}
\rev{
where $\mathrm{sim}(\cdot,\cdot)$ is cosine similarity, and $\tau = 0.1$ is the temperature parameter.
}

\subsubsection{Optimization Objective}
\rev{
During pre-training, for each trajectory $\mathcal T$ in the batch $\mathbb T$, we calculate the average of the reconstruction loss in Equation~\ref{eq:tuple-loss-function} over all generated tuples and add the contrastive loss in Equation~\ref{eq:cl-loss}. Formally:
}
\begin{equation}
    \label{eq:pretraining-loss}
    \rev{
    \mathcal L_{\mathcal T} = \frac{1}{|\mathcal T|}\sum_{i=1}^{|\mathcal T|} \mathcal L_i \;+\; \mathcal L_{\mathcal T}^\mathrm{CL}
    }
\end{equation}

\rev{
Finally, the objective is to minimize:
}
\begin{equation}
    \rev{
    \argmax_{\theta} \sum_{\mathcal T\in \mathbb T} -\mathcal L_{\mathcal T},
    }
\end{equation}
\rev{
where $\theta$ denotes all learnable parameters in the model. By applying this objective across all batches in the training set, we can effectively pre-train UVTM. Algorithm~\ref{alg:pre-training} summarizes the pre-training process.
}

\begin{algorithm}
    \caption{\rev{UVTM Pre-training Epoch}}
    \label{alg:pre-training}
    \begin{algorithmic}[1]
\Require Dense trajectory dataset $\mathcal{D}$, UVTM model $f_\theta$
\For{each batch $\mathbb{T}$ in $\mathcal{D}$}
    \State Initialize an empty embedding set $\mathbb{E}$
    \For{each trajectory $\mathcal{T} \in \mathbb{T}$}
        \State Create a sparse, feature-incomplete version $\mathcal{T}^\mu$ by resampling and feature removal
        \State Reconstruct missing features of $\mathcal{T}^\mu$
        \State Compute reconstruction losses $\mathcal L_i$
        \State Compute embedding $\boldsymbol{O}_{[cls]}^{\mathcal{T}^\mu}$ and add it to $\mathbb{E}$
    \EndFor
    \For{each trajectory $\mathcal{T} \in \mathbb{T}$}
        \State Compute embedding $\boldsymbol{O}_{[cls]}^{\mathcal{T}}$
        \State Compute contrastive loss $\mathcal L_{\mathcal{T}}^\mathrm{CL}$ between $\boldsymbol{O}_{[cls]}^{\mathcal{T}}$ and all embeddings in $\mathbb{E}$
        \State Sum $\mathcal L_i$ and $\mathcal L_{\mathcal{T}}^\mathrm{CL}$ to get $\mathcal L_{\mathcal{T}}$
    \EndFor
    \State Sum $\mathcal L_{\mathcal{T}}$ over the batch to get the batch loss
    \State Update model parameters $\theta$ using the batch loss
\EndFor
\end{algorithmic}
\end{algorithm}

\subsection{Task Adaptation} \label{sec:task-adaptation}
The proposed model’s flexibility allows it to adapt to a variety of trajectory-related tasks after pre-training, particularly when trajectory features are incomplete. We provide four representative tasks as examples.

\subsubsection{Origin-Destination Travel Time Estimation (OD TTE)}
\rev{
We adopt an OD-based variant of the TTE task, which estimates the travel time of a trajectory given only its origin $l_o$, destination $l_d$, and departure time $t_o$. The model receives an input sequence $\langle (l_o, t_o, [m]), (l_d, [m], [m]) \rangle$. It then generates the block corresponding to $(l_d, [m], [m])$ as described in Section~\ref{sec:generation-of-feature-domains}. The generated block is $\langle ((\hat l_d, \Omega(\hat l_d)), \hat t_d, (\hat e_d, \hat r_d)), g_{[e]} \rangle$. Finally, $\hat t_d$ is taken as the estimated arrival time.
}

\subsubsection{Trajectory Recovery (TR)}
\rev{
This task aims to recover a dense version of a sparse trajectory. The procedure is similar to that in Section~\ref{sec:reconstruction-procedure}, except that blocks are generated in their original order without shuffling. Specifically, consider two consecutive points $(l_i, t_i)$ and $(l_{i+1}, t_{i+1})$ in the sparse trajectory, and let $\eta$ be the desired sampling interval for the recovered trajectory. If $t_{i+1} - t_i > \eta$, we insert the tuple $g_{[m]}$ between these two points, indicating a missing sub-trajectory segment to be generated.
}

\subsubsection{Trajectory Prediction (TP)} 
\rev{
This task predicts future trajectory points given historical ones. Suppose the first $n$ points of a trajectory $\mathcal T$ are observed. The model takes the input sequence $\langle g_1^{\mathcal T}, \dots, g_n^{\mathcal T}, g_{[m]} \rangle$, then auto-regressively generates the block of future tuples for $g_{[m]}$, producing $\langle \hat g_{n+1}^\mathcal T, \hat g_{n+2}^\mathcal T, \dots, \hat g_N^\mathcal T, g_{[e]} \rangle$, where $N$ is the trajectory’s total length.
}

\subsubsection{\rev{Similar Search (SS)}}
\rev{
This task identifies which sparse trajectory most closely matches a query dense trajectory from a set of candidates with specific sampling intervals $\mu$. Given a query $\mathcal T$, the model takes $\langle [cls], g_1^{\mathcal T}, g_2^{\mathcal T}, \dots, g_{|\mathcal T|}^{\mathcal T}\rangle$ and produces an embedding $\boldsymbol O_{[cls]}^{\mathcal T}$. For each candidate $\mathcal T'^\mu$, the model similarly obtains $\boldsymbol O_{[cls]}^{\mathcal T'^\mu}$. The candidate with the highest cosine similarity to $\boldsymbol O_{[cls]}^{\mathcal T}$ is selected as the most similar.
}

\rev{
After pre-training, the model can be applied directly (zero-shot) to any of these tasks by feeding in the corresponding input structure. In our experiments, we further fine-tune the model with task-specific data to enhance its performance, except for the SS task which only follows zero-shot. Subsequent sections compare the results of fine-tuned and zero-shot models, examine fine-tuning convergence speed, and investigate scalability with respect to larger fine-tuning datasets.
}

\section{Experiments} \label{sec:experiments}
We evaluate the proposed model on three real-world trajectory datasets, examining its performance under various experimental settings.

\subsection{Datasets}
\rev{Our experiments use three real-world vehicle GPS trajectory datasets: Chengdu, Xian, and Porto. Chengdu and Xian (both in China) record online car-hailing trips recorded by Didi~\footnote{\url{https://gaia.didichuxing.com/}}, while Porto (in Portugal) records taxi trips provided via Kaggle\footnote{\url{https://www.kaggle.com/competitions/pkdd-15-predict-taxi-service-trajectory-i/data}}. We also collect the road networks for these cities from OpenStreetMap\footnote{\url{https://www.openstreetmap.org/}} for map-matching. We use Chengdu and Porto to evaluate the overall performance of each method, and Xian to evaluate the dataset zero-shot performance.}
For consistency, we standardize all trajectories to a 15-second sampling interval. We exclude any trajectory with fewer than 6 points. Table~\ref{tab:dataset-statistic} summarizes the statistics of these datasets.

\begin{table}[h]
    \centering
    \caption{Dataset statistics.}
    \label{tab:dataset-statistic}
    \resizebox{1.0\linewidth}{!}{\begin{tabular}{c|cccc}
  \toprule
  Dataset & Chengdu & \rev{Xian} & Porto \\
  \midrule
  Time span & 10.01--11.30, 2018 & \rev{09.29--10.15, 2018} & 07.01--09.01, 2013 \\
  Longitude & 104.0430$\sim$104.1265 & \rev{108.9168$\sim$109.0043} & -8.6520$\sim$-8.5780 \\
  Latitude & 30.6552$\sim$30.7270 & \rev{34.2062$\sim$34.2803} & 41.1420$\sim$41.1740 \\
  $|\mathcal E|$ & 2,505 & \rev{3,392} & 2,225 \\
  \#Trajectories & 121,394 & \rev{210,000} & 55,120 \\
  \#Points & 3,032,212 & \rev{3,135,991} & 1,482,751 \\
  \bottomrule
\end{tabular}
}
\end{table}

\subsection{Comparison Methods}
To measure our model’s effectiveness on the four tasks in Section~\ref{sec:task-adaptation}, we compare it against various state-of-the-art methods, including task-specific solutions and universal trajectory models.

\subsubsection{OD TTE Methods}
We compare the proposed model’s origin-destination travel time estimation (OD TTE) performance with the following baselines:
\begin{itemize}[leftmargin=*]
    \item \textbf{RNE}~\cite{huang2021learning}: Learns latent embeddings to determine distances among road segments.
    \item \textbf{TEMP}~\cite{wang2019simple}: Uses the historical mean travel times of similar OD pairs.
    \item \textbf{LR}: A linear regression model mapping input features to travel time.
    \item \textbf{GBM}: A non-linear gradient boosting model (XGBoost~\cite{chen2016xgboost}).
    \item \textbf{ST-NN}~\cite{jindal2017unified}: Jointly predicts travel distance and travel time for OD pairs.
    \item \textbf{MURAT}~\cite{li2018multi}: Predicts travel distance and time, incorporating departure time as an additional feature.
    \item \textbf{DeepOD}~\cite{yuan2020effective}: Exploits correlations between input features and historical trajectories.
    \item \textbf{DOT}~\cite{DBLP:journals/corr/abs-2307-03048}: Transforms trajectories into image representations for two-stage travel time estimation.
\end{itemize}

\subsubsection{TR Methods}
We compare the proposed model with the following trajectory recovery (TR) approaches:
\begin{itemize}[leftmargin=*]
    \item \textbf{Shortest Path}~\cite{DBLP:journals/tsas/ChambersFWW20}: Uses Dijkstra’s algorithm~\cite{johnson1973note} to find paths between consecutive points in sparse trajectories.
    \item \textbf{Linear}~\cite{DBLP:journals/cn/HoteitSSRP14}: Recovers trajectories via linear interpolation.
    \item \textbf{TrImpute}~\cite{DBLP:conf/gis/ElshrifIM22}: Imputes sparse trajectories based on crowd wisdom.
    \item \textbf{DHTR}~\cite{wang2019deep}: Combines a seq2seq framework~\cite{DBLP:conf/nips/SutskeverVL14} with Kalman filtering to recover dense trajectories.
    \item \textbf{AttnMove}~\cite{DBLP:conf/aaai/XiaQ0XSGL21}: Uses attention mechanisms to predict road segments in a sequence.
    \item \textbf{MTrajRec}~\cite{DBLP:conf/kdd/RenRL0ML021}: A GRU-based auto-regressive model that recovers trajectories constrained by road networks.
    \item \textbf{RNTrajRec}~\cite{DBLP:journals/corr/abs-2211-13234}: Incorporates transformers and road network structure to recover dense trajectories.
\end{itemize}

\subsubsection{TP Methods}
The following methods are specifically designed for the trajectory prediction (TP) task:
\begin{itemize}[leftmargin=*]
    \item \textbf{DeepMove}~\cite{feng2018deepmove}: An attentional RNN-based approach that predicts future movement.
    \item \textbf{Transformer}~\cite{DBLP:conf/nips/VaswaniSPUJGKP17}: A powerful sequence model that captures long-range dependencies.
    \item \rev{\textbf{PreCLN}~\cite{yan2022precln}: Improves trajectory prediction accuracy by integrating contrastive learning.}
\end{itemize}

\subsubsection{\rev{Universal Trajectory Models}}
We also compare our model to universal trajectory models on the TP and SS tasks:
\begin{itemize}[leftmargin=*]
    \item \textbf{trajectory2vec}~\cite{DBLP:conf/ijcnn/YaoZZHB17}: Uses behavior sequences to capture high-level correlations in trajectories.
    \item \textbf{t2vec}~\cite{DBLP:conf/icde/LiZCJW18}: Builds upon an auto-encoding strategy for a more general trajectory representation.
    \item \textbf{Trembr}~\cite{fu2020trembr}: Integrates road network information within an auto-encoder to learn generic trajectory embeddings.
    \item \textbf{START}~\cite{jiang2022self}: Incorporates both spatiotemporal correlations and travel semantics for universal trajectory modeling.
    \item \rev{\textbf{PathLLM}~\cite{wei2025path}: Uses large language models for multi-modal path representation learning.}
    \item \rev{\textbf{PLM4Traj}~\cite{zhou2024plm4traj}: A multi-task trajectory model that builds on pre-trained language models.}
    \item \rev{\textbf{BIGCity}~\cite{yu2024bigcity}: A multi-task trajectory model that unifies trajectory and traffic state data with large language models.}
\end{itemize}

Since \textit{Trembr}, \textit{START}, \rev{\textit{PathLLM}, and \textit{BIGCity}} rely on road-segment features, which are not directly available for sparse trajectories, we first use RNTrajRec (the best-performing TR baseline) to recover missing segments and then feed them into these models. We denote this process as \textit{Trembr+RNTR}, \textit{START+RNTR}, \rev{\textit{PathLLM+RNTR} and \textit{BIGCity+RNTR}}. We also report each model's performance when map-matched real dense trajectories are available.

\subsubsection{Variations Without Pre-training or Fine-tuning}
To further evaluate the effectiveness of pre-training and fine-tuning, we introduce two model variations:
\begin{itemize}[leftmargin=*]
    \item \textbf{UVTM w/o pt}: Trains the model with task-specific generation alone, omitting the pre-training in Section~\ref{sec:pre-training}.
    \item \textbf{UVTM w/o ft}: Uses the pre-trained model directly in a zero-shot setting, bypassing any task-specific fine-tuning.
\end{itemize}

\subsection{Settings}
We sort all trajectories by their departure time in each dataset, then split them into training, validation, and testing sets with an 8:1:1 ratio. All models are trained on the training set, and we use the validation set for hyper-parameter tuning and early stopping. The final metrics are reported on the testing set.
For the TR, TP, and SS tasks, we simulate different sparsity levels by using trajectories sampled at 1-, 2-, and 4-minute intervals. The TR task recovers these sparse trajectories back to a 15-second interval. The TP task predicts the road segments, fraction, coordinates, and times of trajectory destinations, given the entire historical trajectory except the final point. \rev{The SS task identifies the most similar sparse trajectory from a set of candidate trajectories, with a dense trajectory serving as the query. The candidate set derives from all testing-set trajectories.}

We use different metrics to evaluate each downstream task:
\textit{1)~OD TTE:} We measure root mean squared error (RMSE), mean absolute error (MAE), and mean absolute percentage error (MAPE) to assess travel time estimation accuracy.
\textit{2)~TR:} Following MTrajRec~\cite{DBLP:conf/kdd/RenRL0ML021}, we compute the Precision and Recall of the recovered road segments. For each sample, let $\mathcal E_R$ be the set of recovered road segments and $\mathcal E_G$ the set of ground-truth segments. We define $\mathrm{Precision} = \frac{|\mathcal E_R \cap \mathcal E_G|}{|\mathcal E_R|}$ and $\mathrm{Recall} = \frac{|\mathcal E_R \cap \mathcal E_G|}{|\mathcal E_G|}$. We also measure the MAE of the recovered coordinates and road locations. The distance between recovered and ground-truth points is calculated either on the Earth’s surface or the road network, depending on whether we evaluate recovered coordinates or road locations.
\textit{3)~TP:} We report the Accuracy of the predicted road-segment indices, the distance MAE of the predicted coordinates and road locations, and the MAE of the predicted time.
\rev{\textit{4)~SS:} We report the Mean Rank and Accuracy of the identified most similar trajectories.}
Each experiment is conducted 5 times and the average metric values are reported.

The proposed method is implemented using Python and PyTorch~\cite{paszke2019pytorch}\footnote{Codes available at \url{https://github.com/Logan-Lin/UVTM}.}. Baselines are configured with the optimal parameters recommended in their respective papers. \rev{We tune the key hyper-parameters of our model over preset ranges on the validation set of Chengdu, adjusting one hyper-parameter at a time while fixing others. The optimal values are listed in Table~\ref{tab:optimal-parameter}.}

\begin{table}[h]
    \centering
    \caption{Hyper-parameter range and optimal values.}
    \label{tab:optimal-parameter}
    \resizebox{1.0\linewidth}{!}{\begin{tabular}{c|cc}
    \toprule
    Parameter & \rev{Description} & Range \& Optimal Value \\
    \midrule
    $d$ & \rev{Embedding dimension} & 32, 64, \underline{128}, 192, 256 \\
    $N_h$ & \rev{Number of Attention heads} & 1, 2, 4, \underline{8}, 16 \\
    $\delta$ & \rev{Distance threshold in meters} & 10, 50, \underline{100}, 150, 200 \\
    \rev{$\varphi$} & \rev{Feature removal probability} & \rev{0\%, 10\%, \underline{20\%}, 30\%, 40\%} \\
    \rev{$|\mathbb T|$} & \rev{Batch size during pre-training} & \rev{32, 64, \underline{128}, 256, 512} \\
    \bottomrule
\end{tabular}}
\end{table}

\subsection{Comparison with Baselines}

\begin{table}
    \centering
    \caption{Origin-destination travel time estimation accuracy of different approaches.}
    \label{tab:travel-time-estimation-accuracy}
    \begin{threeparttable}
\resizebox{1.0\linewidth}{!}{
\begin{tabular}{c|ccc}
\toprule
Datasets & \multicolumn{3}{c}{Chengdu / Porto} \\
\midrule
Methods & MAE (min) $\downarrow$ & RMSE (min) $\downarrow$ & MAPE (\%) $\downarrow$ \\

\midrule
RNE & 1.087 / 2.357 & 4.967 / 7.168 & 18.185 / 53.894 \\
TEMP & 0.816 / 2.610 & 1.100 / 3.414 & 13.003 / 59.178 \\
LR & 0.815 / 2.596 & 1.097 / 3.408 & 12.997 / 58.390 \\
GBM & 0.773 / 2.200 & 1.202 / 3.116 & 11.142 / 43.308 \\
ST-NN & 0.770 / 2.136 & 1.031 / 3.027 & 12.470 / 45.285 \\
MURAT & 0.731 / 1.971 & 0.979 / 2.827 & 11.931 / 41.259 \\
DeepOD & 0.640 / 1.899 & 0.880 / 2.780 & 10.517 / 36.956 \\
DOT & \underline{0.614} / \underline{1.777} & \underline{0.841} / \underline{2.644} & \underline{9.937} / \underline{34.883} \\
\midrule
UVTM w/o pt & 0.666 / 1.871 & 0.933 / 2.797 & 10.501 / 34.895 \\
UVTM w/o ft & \rev{0.725 / 2.225} & \rev{0.963 / 2.981} & \rev{11.148 / 40.615} \\
\textbf{UVTM} & \rev{\textbf{0.558} / \textbf{1.602}} & \rev{\textbf{0.779} / \textbf{2.450}} & \rev{\textbf{8.745} / \textbf{31.270}} \\
\bottomrule
\end{tabular}
}
\begin{tablenotes}\footnotesize
\item[]{
    \textbf{Bold} denotes the best result, \underline{underline} denotes the second-best result.
    $\downarrow$ means lower is better.
}
\end{tablenotes}
\end{threeparttable}
\end{table}

\begin{table*}
    \caption{Trajectory recovery accuracy of different approaches.}
    \label{tab:sparse-trajectory-recovery-accuracy}
    \begin{threeparttable}
\resizebox{1.0\linewidth}{!}{
\begin{tabular}{c|c|cccc}
\toprule
\multicolumn{2}{c|}{Sampling Interval $\mu$} & \multicolumn{4}{c}{1 minute / 2 minutes / 4 minutes} \\
\midrule
Datasets & Methods & Precision (\%) $\uparrow$ & Recall (\%) $\uparrow$ & MAE (Coor, meters) $\downarrow$ & MAE (Road, meters) $\downarrow$ \\

\midrule
\multirow{10}{*}{Chengdu}
& Shortest Path & 62.638 / 43.504 / 29.431 & 59.346 / 40.949 / 27.607 & 213.10 / 428.69 / 752.19 & 206.91 / 391.39 / 586.09 \\
& Linear & 66.642 / 48.604 / 36.209 & 65.557 / 45.234 / 30.496 & 183.64 / 385.23 / 675.85 & 169.46 / 378.96 / 564.19 \\
& TrImpute & 77.520 / 60.179 / 57.526 & 76.202 / 58.461 / 53.747 & 166.82 / 276.56 / 408.64 & 155.71 / 265.34 / 387.36 \\
& DHTR & 53.514 / 53.608 / 50.985 & 58.868 / 47.918 / 46.311 & 205.59 / 317.45 / 450.61  & 300.67 / 470.46 / 547.41 \\
& AttnMove & 84.162 / 81.402 / 78.645 & 81.839 / 76.612 / 69.257 & 252.59 / 280.20 / 354.39 & 201.51 / 258.69 / 323.52 \\
& MTrajRec & 85.039 / 82.596 / \underline{80.684} & 83.351 / 80.113 / \underline{72.952} & 243.01 / 264.15 / \underline{311.53} & 173.67 / 204.58 / \underline{282.88} \\
& RNTrajRec & 87.653 / 83.174 / 79.404 & \underline{86.025} / \underline{80.150} / 72.633 & 215.24 / 234.27 / 326.92 & 114.04 / \underline{148.04} / 292.61 \\
\cmidrule{2-6}
& UVTM w/o pt & 83.720 / 77.425 / 72.471 & 82.827 / 73.933 / 62.757 & 194.30 / 272.86 / 479.49 & 86.15 / 230.68 / 448.98 \\
& UVTM w/o ft
  & \rev{\underline{87.658} / \underline{83.605} / 79.102}
  & \rev{85.845 / 78.412 / 70.716}
  & \rev{\underline{137.82} / \underline{218.33} / 350.51}
  & \rev{\underline{82.13} / 167.11 / 315.62} \\
& \textbf{UVTM}
  & \rev{\textbf{89.064} / \textbf{84.361} / \textbf{80.839}}
  & \rev{\textbf{88.240} / \textbf{81.518} / \textbf{73.210}}
  & \rev{\textbf{133.46} / \textbf{192.66} / \textbf{305.08}}
  & \rev{\textbf{68.01} / \textbf{143.55} / \textbf{275.49}} \\

\midrule
\multirow{10}{*}{Porto}
& Shortest Path & 69.780 / 53.590 / 40.492 & 60.354 / 46.263 / 33.758 & 202.17 / 434.37 / 679.72 & 165.02 / 319.32 / 478.66 \\
& Linear & 72.961 / 60.966 / 48.529 & 63.146 / 48.401 / 35.507 & 196.51 / 403.05 / 621.88 & 132.76 / 275.43 / 430.22 \\
& TrImpute & 76.781 / 66.492 / 50.021 & 69.599 / 58.676 / 43.052 & 132.76 / 275.43 / 430.22 & 128.48 / 235.63 / 347.30 \\
& DHTR & 63.287 / 58.897 / 52.658 & 62.511 / 56.444 / 42.462 & 235.32 / 292.65 / 355.23 & 285.68 / 336.48 / 389.18 \\
& AttnMove & 79.541 / 75.751 / 71.248 & 67.116 / 56.751 / 48.991 & 184.70 / 222.51 / 304.31 & 134.17 / 184.03 / 251.92 \\
& MTrajRec & 78.081 / 72.847 / 64.566 & 71.853 / 60.068 / 46.110 & 168.34 / 283.90 / 496.96 & 121.64 / 215.40 / 391.67 \\
& RNTrajRec & 80.305 / 77.094 / \underline{75.573} & 74.953 / \underline{65.370} / \underline{50.965} & 135.17 / \underline{175.42} / \underline{294.08} & 111.75 / 152.49 / \textbf{230.30} \\
\cmidrule{2-6}
& UVTM w/o pt & 81.058 / 73.842 / 67.189 & 77.683 / 61.987 / 42.688 & 132.97 / 229.11 / 496.99 & \underline{69.70} / 204.01 / 488.13 \\
& UVTM w/o ft
  & \rev{\underline{81.708} / \underline{79.073} / 74.727}
  & \rev{\underline{75.138} / 63.057 / 50.275}
  & \rev{\underline{118.82} / 194.43 / 338.46}
  & \rev{76.52 / \underline{159.85} / 320.91} \\
& \textbf{UVTM}
  & \rev{\textbf{82.645} / \textbf{79.501} / \textbf{77.073}}
  & \rev{\textbf{78.607} / \textbf{66.859} / \textbf{53.439}}
  & \rev{\textbf{103.03} / \textbf{164.82} / \textbf{282.69}}
  & \rev{\textbf{63.22} / \textbf{138.74} / \underline{242.11}} \\
\bottomrule
\end{tabular}
}
\begin{tablenotes}\footnotesize
\item[]{
    \textbf{Bold} denotes the best result, and \underline{underline} denotes the second-best result.
    $\uparrow$ means higher is better, and $\downarrow$ means lower is better.
}
\end{tablenotes}
\end{threeparttable}
\end{table*}

\begin{table*}
    \caption{Trajectory prediction accuracy of different approaches.}
    \label{tab:trajectory-prediction-accuracy}
    \begin{threeparttable}
\resizebox{1.0\linewidth}{!}{
\begin{tabular}{c|c|cccc}
\toprule
\multicolumn{2}{c|}{Sampling Interval $\mu$} & \multicolumn{4}{c}{1 minute / 2 minutes / 4 minutes} \\
\midrule
Datasets & Methods & Accuracy (\%) $\uparrow$ & MAE (Coor, meters) $\downarrow$ & MAE (Road, meters) $\downarrow$ & MAE (Time, seconds) $\downarrow$ \\

\midrule
\multirow{13}{*}{Chengdu}
& trajectory2vec & 31.496 / 24.403 / 17.163 & 1514.5 / 1682.8 / 1861.6 & 1322.0 / 1616.5 / 1957.5 & 14.474 / 20.722 / 37.256 \\
& t2vec & 53.349 / 43.303 / 35.058 & 528.65 / 602.45 / 731.00 & 286.18 / 434.27 / 635.45 & 13.016 / 19.539 / 34.488 \\
& DeepMove & 58.499 / 45.985 / 37.338 & \underline{319.14} / 461.73 / 664.93 & 258.96 / 397.95 / 607.42 & 11.994 / 19.435 / 35.039 \\
& Transformer & 65.192 / 60.028 / 55.139 & 374.36 / 402.11 / 431.01 & 236.86 / 287.61 / 320.92 & 16.287 / 29.848 / 34.226 \\
& \rev{PreCLN}           & \rev{66.736 / 61.442 / 55.200}
                    & \rev{365.12 / 395.83 / 427.22}
                    & \rev{225.15 / 278.62 / 316.11}
                    & \rev{15.462 / 28.057 / 33.154} \\
& Trembr+RNTR & 52.065 / 43.196 / 34.655 & 421.95 / 482.67 / 561.89 & 398.76 / 455.98 / 532.03 & 14.346 / 19.110 / 28.659 \\
& START+RNTR & 59.462 / 48.466 / 40.941 & 375.39 / 421.45 / 481.32 & 355.00 / 399.66 / 457.35 & 12.771 / 14.439 / 19.443 \\
& \rev{PathLLM+RNTR}     & \rev{66.356 / 58.319 / 49.884}
                    & \rev{370.55 / 417.83 / 478.51}
                    & \rev{345.76 / 386.33 / 449.29}
                    & \rev{12.025 / 13.833 / 18.986} \\
& \rev{PLM4Traj}         & \rev{\underline{73.458} / \underline{66.815} / \underline{58.463}}
                    & \rev{320.47 / \underline{355.91} / \underline{405.67}}
                    & \rev{209.36 / \underline{269.48} / \underline{305.14}}
                    & \rev{13.865 / 24.199 / 31.007} \\
& \rev{BIGCity+RNTR}     & \rev{68.229 / 60.044 / 50.036}
                    & \rev{368.17 / 415.28 / 476.92}
                    & \rev{340.41 / 382.05 / 446.98}
                    & \rev{11.896 / 13.547 / 18.762} \\
\cmidrule{2-6}
& UVTM w/o pt & 71.795 / 50.041 / 33.882 & 376.05 / 540.89 / 857.43 & 263.83 / 568.73 / 942.64 & \underline{5.301} / \underline{10.198} / 9.562 \\
& UVTM w/o ft
             & \rev{68.912 / 63.273 / 55.011}
             & \rev{324.34 / 390.73 / 463.14}
             & \rev{\underline{202.91} / 279.18 / 382.98}
             & \rev{5.734 / 11.453 / \underline{9.207}} \\
& \textbf{UVTM}
             & \rev{\textbf{83.052} / \textbf{79.236} / \textbf{72.560}}
             & \rev{\textbf{257.19} / \textbf{297.07} / \textbf{352.54}}
             & \rev{\textbf{126.14} / \textbf{197.62} / \textbf{253.46}}
             & \rev{\textbf{3.769} / \textbf{7.855} / \textbf{7.094}} \\

\midrule
\multirow{13}{*}{Porto}
& trajectory2vec & 13.396 / 10.178 / 5.440 & 1709.3 / 2108.9 / 2488.5 & 2227.4 / 2425.6 / 3003.6 & 31.442 / 46.097 / 54.585  \\
& t2vec & 38.945 / 30.805 / 22.812 & 432.77 / 528.89 / 732.13 & 206.95 / 346.72 / 641.26 & 17.420 / 28.436 / 48.651 \\
& DeepMove & 43.774 / 33.562 / 23.645 & \underline{252.22} / 390.57 / 679.62 & 197.89 / 328.93 / 681.40 & 16.998 / 26.309 / 46.629 \\
& Transformer & 43.441 / 39.425 / 33.685 & 323.58 / 351.18 / 402.32 & 216.18 / 256.89 / 283.72  & 18.231 / 30.541 / 49.083 \\ 
& \rev{PreCLN}           & \rev{45.329 / 41.768 / 36.582}
                    & \rev{310.45 / 340.59 / 395.02}
                    & \rev{210.87 / 250.22 / 280.44}
                    & \rev{17.456 / 29.087 / 47.652} \\
& Trembr+RNTR & 40.128 / 34.857 / 26.004 & 413.20 / 471.45 / 620.77 & 393.18 / 448.74 / 597.73 & 18.915 / 21.843 / 28.683 \\
& START+RNTR & 52.118 / 43.617 / 34.931 & 351.98 / 416.83 / 503.32 & 333.56 / 396.39 / 483.41 & 14.729 / 17.722 / 22.455 \\
& \rev{PathLLM+RNTR}         & \rev{55.213 / 48.836 / 38.527}
                    & \rev{341.93 / 411.50 / 498.12}
                    & \rev{324.54 / 390.73 / 472.66}
                    & \rev{13.701 / 20.438 / 27.516} \\
& \rev{PLM4Traj}     & \rev{58.341 / 53.846 / 50.091}
                    & \rev{298.12 / \underline{329.45} / \underline{380.57}}
                    & \rev{204.33 / 240.56 / 275.11}
                    & \rev{11.345 / 14.261 / 18.663} \\
& \rev{BIGCity+RNTR}     & \rev{57.089 / 52.351 / 41.293}
                    & \rev{338.64 / 408.12 / 495.77}
                    & \rev{320.03 / 386.67 / 468.92}
                    & \rev{10.917 / 13.358 / 18.035} \\
\cmidrule{2-6}
& UVTM w/o pt & \underline{61.480} / 46.938 / 31.304 & 307.23 / 369.86 / 649.71 & \underline{145.36} / 295.58 / 624.11 & \underline{10.853} / \underline{11.316} / \underline{11.643} \\
& UVTM w/o ft 
  & \rev{59.100 / \underline{57.650} / \underline{53.130}}
  & \rev{317.12 / 342.74 / 393.67}
  & \rev{166.09 / \underline{180.02} / \underline{242.08}}
  & \rev{11.223 / 15.032 / 15.699} \\
& \textbf{UVTM}
  & \rev{\textbf{68.592} / \textbf{67.671} / \textbf{65.331}}
  & \rev{\textbf{235.73} / \textbf{234.47} / \textbf{314.24}}
  & \rev{\textbf{98.42} / \textbf{104.81} / \textbf{136.88}}
  & \rev{\textbf{6.761} / \textbf{9.547} / \textbf{11.264}} \\
\bottomrule
\end{tabular}
}
\begin{tablenotes}\footnotesize
\item[]{
    \textbf{Bold} denotes the best result, and \underline{underline} denotes the second-best result.
    $\uparrow$ means higher is better, and $\downarrow$ means lower is better.
}
\end{tablenotes}
\end{threeparttable}
\end{table*}

\begin{table}
    \centering
    \caption{\rev{Similarity search accuracy of different approaches.}}
    \label{tab:similar-search-accuracy}
    \resizebox{1.0\linewidth}{!}{\begin{threeparttable}
\begin{tabular}{c|c|cc}
\toprule
\multicolumn{2}{c|}{Sampling Interval $\mu$} & \multicolumn{2}{c}{1 minute / 2 minutes / 4 minutes} \\
\midrule
Datasets & Methods & Mean Rank $\downarrow$ & Accuracy (\%) $\uparrow$ \\

\midrule
\multirow{8}{*}{Chengdu}
& \rev{trajectory2vec}    & 
  \rev{3.409 / 3.736 / 4.250}
  & \rev{81.99 / 75.00 / 67.99} \\

& \rev{t2vec}             & 
  \rev{3.384 / 3.610 / 4.120}
  & \rev{80.19 / 75.72 / 68.67} \\

& \rev{Trembr+RNTR}       & 
  \rev{5.395 / 5.689 / 6.114}
  & \rev{68.77 / 60.03 / 52.97} \\

& \rev{START+RNTR}        & 
  \rev{3.089 / 3.405 / 3.888}
  & \rev{72.21 / 64.66 / 57.90} \\

& \rev{PathLLM+RNTR}      & 
  \rev{5.280 / 5.610 / 6.001}
  & \rev{69.14 / 60.91 / 53.49} \\

& \rev{PLM4Traj}          &
  \rev{\textbf{1.079} / \underline{1.222} / \underline{1.356}}
  & \rev{\textbf{98.03} / \underline{86.21} / \underline{75.51}} \\

& \rev{BIGCity+RNTR}      &
  \rev{2.501 / 2.922 / 3.510}
  & \rev{76.72 / 69.81 / 62.77} \\

& \rev{\textbf{UVTM w/o ft}} & \rev{\underline{1.081} / \textbf{1.193} / \textbf{1.299}} & \rev{\underline{97.99} / \textbf{88.82} / \textbf{80.70}} \\

\midrule
\multirow{8}{*}{Porto}
& \rev{trajectory2vec}    & 
  \rev{3.559 / 3.943 / 4.298}
  & \rev{80.01 / 73.99 / 66.59} \\

& \rev{t2vec}             &
  \rev{3.488 / 3.799 / 4.013}
  & \rev{80.87 / 74.64 / 67.03} \\

& \rev{Trembr+RNTR}       &
  \rev{3.203 / 3.628 / 3.973}
  & \rev{75.48 / 68.35 / 60.92} \\

& \rev{START+RNTR}        &
  \rev{2.611 / 3.027 / 3.438}
  & \rev{78.66 / 71.84 / 64.11} \\

& \rev{PathLLM+RNTR}      &
  \rev{3.085 / 3.483 / 3.810}
  & \rev{76.21 / 69.26 / 61.45} \\

& \rev{PLM4Traj}          &
  \rev{\underline{1.146} / \underline{1.263} / \underline{1.397}}
  & \rev{\underline{97.60} / \underline{85.33} / \underline{73.77}} \\

& \rev{BIGCity+RNTR}      &
  \rev{2.278 / 2.798 / 3.243}
  & \rev{81.90 / 75.03 / 68.08} \\
& \rev{\textbf{UVTM w/o ft}} & \rev{\textbf{1.047} / \textbf{1.155} / \textbf{1.257}} & \rev{\textbf{98.42} / \textbf{89.17} / \textbf{81.24}} \\
\bottomrule
\end{tabular}
\begin{tablenotes}\footnotesize
\item[]{
    \textbf{Bold} denotes the best result, \underline{underline} denotes the second-best result.
    $\uparrow$ means higher is better, and $\downarrow$ means lower is better.
}
\end{tablenotes}
\end{threeparttable}}
\end{table}

\begin{table}
    \centering
    \caption{Trajectory prediction accuracy comparison between baselines using dense trajectories and the proposed model.}
    \label{tab:dense-trajectory-prediction-accuracy}
    \resizebox{1.0\linewidth}{!}{\begin{threeparttable}
\begin{tabular}{cc|ccc}
\toprule
\multicolumn{2}{c|}{Datasets} & \multicolumn{3}{c}{Chengdu / Porto} \\
\midrule
Methods & $\mu$ & Accuracy (\%) & MAE (Coor, m.) & MAE (Time, sec.) \\

\midrule
Trembr & 15 sec. & 66.077 / 52.063 & 399.40 / 362.84 & 8.534 / 15.046 \\
START & 15 sec. & 74.990 / 62.997 & 354.76 / 318.26 & 8.068 / 12.476 \\
\rev{PathLLM}  & \rev{15 sec.} & \rev{74.612 / 63.129} & \rev{348.51 / 311.09} & \rev{7.846 / 11.674} \\
\rev{BIGCITY}  & \rev{15 sec.} & \rev{\underline{78.442} / \underline{65.515}} & \rev{\underline{325.39} / \underline{299.87}} & \rev{7.724 / \underline{11.098}} \\
\midrule
\textbf{UVTM} & \textbf{4 min.} & \rev{72.560 / 65.331} & \rev{352.54 / 314.24} & \rev{\underline{7.094} / 11.264} \\
\textbf{UVTM} & \textbf{1 min.} & \rev{\textbf{83.052} / \textbf{68.592}} & \rev{\textbf{257.19} / \textbf{235.73}} & \rev{\textbf{3.769} / \textbf{6.761}} \\
\bottomrule
\end{tabular}
\begin{tablenotes} \footnotesize
    \item[]{\textbf{Bold} denotes the best result, \underline{underline} denotes the second-best result.}
\end{tablenotes}
\end{threeparttable}}
\end{table}

\subsubsection{Comparison on Overall Accuracy}
Tables~\ref{tab:travel-time-estimation-accuracy} to \ref{tab:similar-search-accuracy} show the performance of various methods on OD TTE, TR, TP, and SS tasks, respectively. The proposed model achieves consistently strong results across these tasks, highlighting its adaptability.

On the OD TTE task, the proposed model surpasses \textit{DOT}, the strongest baseline in this task. Its advantage largely stems from pre-training, which helps the model better capture the relationship between trajectories and travel time. This is made obvious by the fact that while \textit{DOT} outperforms \textit{UVTM w/o pt}, it still lags behind the fully trained \textit{UVTM}.

On the TR task, most existing methods train specifically on sparse trajectories with a single sampling interval, limiting their flexibility. By contrast, our model’s pre-training allows it to handle multiple intervals without retraining. Even without fine-tuning (\textit{UVTM w/o ft}), the model matches or outperforms the best baselines, demonstrating its robustness across different sampling intervals.

\rev{
On the TP and SS tasks, although the latest models, such as \textit{PathLLM} and \textit{BIGCity}, perform well on dense trajectories, their accuracy drops significantly for sparse inputs. Pairing them with TR methods introduces error accumulation. Our model, on the other hand, directly accommodates sparse trajectories, leading to more robust performance. Table~\ref{tab:dense-trajectory-prediction-accuracy} further shows that with only 1-minute sampling, it can outperform \textit{BIGCity} trained on dense trajectories sampled at 15-second intervals.
}

\subsubsection{\rev{Comparison on Dataset Zero-shot}}

\begin{table}
    \centering
    \caption{\rev{Origin-destination travel time estimation accuracy comparison of different approaches with the dataset zero-shot setting.}}
    \label{tab:dataset-zero-shot}
    \resizebox{1.0\linewidth}{!}{\begin{threeparttable}
\begin{tabular}{c|ccc}
\toprule
Datasets & \multicolumn{3}{c}{\rev{Chengdu$\rightarrow$Xian / Xian$\rightarrow$Chengdu}} \\
\midrule
Methods & MAE (min) $\downarrow$ & RMSE (min) $\downarrow$ & MAPE (\%) $\downarrow$ \\

\midrule
RNE          & \rev{3.913 / 1.556} & \rev{13.411 / 4.659} & \rev{34.223 / 29.103} \\
TEMP         & \rev{2.955 / 1.737} & \rev{2.992 / 2.238}  & \rev{24.076 / 31.364} \\
LR           & \rev{2.928 / 1.713} & \rev{2.952 / 2.250}  & \rev{24.044 / 30.947} \\
GBM          & \rev{2.809 / 1.467} & \rev{3.259 / 2.025}  & \rev{21.039 / 23.386} \\
ST-NN        & \rev{2.780 / 1.410} & \rev{2.763 / 1.968}  & \rev{23.294 / 24.001} \\
MURAT        & \rev{2.617 / 1.322} & \rev{2.636 / 1.866}  & \rev{22.450 / 22.280} \\
DeepOD       & \rev{2.304 / 1.262} & \rev{2.376 / 1.807}  & \rev{19.268 / 19.960} \\
DOT          & \rev{\underline{2.205} / \underline{1.174}} & \rev{\underline{2.257} / \underline{1.719}}  & \rev{\underline{18.700} / \underline{18.487}} \\
\textbf{UVTM w/o ft} 
             & \rev{\textbf{2.016 / 1.169}}
             & \rev{\textbf{2.101 / 1.698}}
             & \rev{\textbf{16.420 / 17.864}} \\
\bottomrule
\end{tabular}
\begin{tablenotes}\footnotesize
\item[]{
    \textbf{Bold} denotes the best result, \underline{underline} denotes the second-best result.
    $\downarrow$ means lower is better.
    A$\rightarrow$B means each method is trained on dataset A and evaluated on dataset B.
}
\end{tablenotes}
\end{threeparttable}}
\end{table}

\rev{
We also evaluate a dataset zero-shot scenario, where models are trained on one dataset and tested on another without fine-tuning. Table~\ref{tab:dataset-zero-shot} shows results for OD TTE. Compared with Table~\ref{tab:travel-time-estimation-accuracy}, all methods suffer from distribution shifts; however, our proposed model still outperforms the baselines, demonstrating strong generalizability to unseen datasets.
}

\subsubsection{Comparison on Efficiency}

\begin{table}
    \centering
    \caption{Efficiency metrics of different approaches.}
    \label{tab:efficiency}
    \resizebox{1.0\linewidth}{!}{\begin{tabular}{c|c|ccc}
\toprule
\multicolumn{2}{c|}{Datasets} & \multicolumn{3}{c}{Chengdu / Porto} \\
\midrule
\multirow{2}{*}{Tasks} & \multirow{2}{*}{Methods} & Model size & Train time & Test time \\
& & (MBytes) & (min/epoch) & (sec) \\

\midrule
\multirow{6}{*}{OD TTE}
& RNE & 2.446 / 2.173 & 0.100 / 0.040 & 0.170 / 0.062 \\
& ST-NN & 1.185 / 1.185  & 0.112 / 0.082 & 0.220 / 0.085  \\
& MURAT & 9.120 / 8.847 & 0.153 / 0.095  & 0.210 / 0.075  \\
& DeepOD & 8.184 / 7.928 & 0.382 / 0.171  & 0.328 / 0.099  \\
& DOT & 8.763 / 8.496  & 1.552 / 0.752 & 1.672 / 0.926  \\
& \textbf{UVTM} & 10.146 / 9.333 & 0.533 / 0.336 & 1.231 / 0.647 \\

\midrule
\multirow{6}{*}{TR}
& TrImpute & 2.778 / 1.262 & - / - & 6.110K / 2.199K \\
& DHTR & 6.426 / 6.426 & 0.176 / 0.115 & 2.503K / 0.257K \\
& AttnMove & 6.799 / 6.250 & 6.844 / 2.460 & 92.673 / 32.081 \\
& MTrajRec & 19.180 / 18.495 & 8.470 / 4.437 & 0.125K / 54.062 \\
& RNTrajRec & 20.639 / 19.876 & 8.925 / 4.463 & 0.144K / 60.861 \\
& \textbf{UVTM} & 10.146 / 9.333 & 1.668 / 1.470 & 21.984 / 15.237 \\

\midrule
\multirow{11}{*}{TP}
& trajectory2vec & 6.306 / 6.306 & 0.158 / 0.085 & 0.204 / 0.110 \\
& t2vec & 7.170 / 7.170 & 0.278 / 0.109 & 0.221 / 0.113  \\
& DeepMove & 5.282 / 5.282 & 0.253 / 0.126 & 0.567 / 0.292 \\
& Transformer & 6.295 / 6.295 & 1.090 / 0.538 & 3.568 / 1.596 \\
& \rev{PreCLN} & \rev{8.509 / 7.814} & \rev{1.431 / 0.970} & \rev{4.968 / 1.833} \\
& Trembr+RNTR & 26.103 / 25.792 & 9.468 / 5.072 & 0.149K / 62.500 \\
& START+RNTR & 28.708 / 27.099 & 10.198 / 5.420 & 0.158K / 65.468 \\
& \rev{PathLLM+RNTR} & \rev{0.446K / 0.438K} & \rev{41.866 / 28.020} & \rev{0.264K / 0.152K} \\
& \rev{PLM4Traj} & \rev{0.429K / 0.425K} & \rev{24.933 / 17.936} & \rev{0.110K / 85.857} \\
& \rev{BIGCity+RNTR} & \rev{0.507K / 0.487K} & \rev{57.828 / 39.265} & \rev{0.299K / 0.171K} \\
& \textbf{UVTM} & 10.146 / 9.333 & 1.667 / 1.450 & 3.357 / 1.703 \\

\midrule
\rev{Pre-train} & \rev{\textbf{UVTM}} & \rev{10.146 / 9.333} & \rev{1.713 / 1.545} & \rev{- / -} \\

\bottomrule
\end{tabular}}
\end{table}

We evaluate each approach in terms of model size, training time, and testing time. Model size measures storage usage, while training and testing times reflect computational efficiency. We compute model sizes by summing up the learnable parameters of each method. We measure training and testing times on a machine with an Intel(R) Xeon(R) Gold 5215 CPU and an NVIDIA(R) Quadro RTX 8000 GPU.

Table~\ref{tab:efficiency} shows that the proposed model achieves efficiency comparable to or better than the leading task-specific methods. \rev{We also report the pre-training efficiency of our model in the last row of the table.} Moreover, in real-world scenarios where multiple tasks must be performed on the same dataset, our model becomes even more efficient. Since it can handle different tasks after a single training procedure, it requires less computational time and storage compared to maintaining multiple task-specific models. \rev{For example, to run TTE, TR, and TP on the Chengdu dataset, choosing DOT, RNTrajRec, and START for these tasks would take 20.675 minutes per training epoch, while pre-training our model takes only 1.713 minutes per epoch.}

\subsection{Performance Analysis}

\subsubsection{Efficacy of Pre-training}
We examine how pre-training influences the model’s convergence rate on each task, comparing versions with and without pre-training. Specifically, we track performance metrics over training epochs. Figure~\ref{fig:convergence-rate} displays the results.

Across both TR and TP tasks, the pre-trained model converges faster than the model trained from scratch. This indicates that pre-training helps the model adapt to new tasks with fewer epochs. As a result, pre-training provides computational savings, which is especially beneficial when the same dataset is used for multiple tasks.

\begin{figure}
    \centering
    \input{plots/convergence}
    \caption{Comparison of convergence rate with or without pre-training in Chengdu.}
    \label{fig:convergence-rate}
\end{figure}

\subsubsection{Scalability of Pre-training with Limited Data}
Our approach requires a certain amount of dense trajectories for pre-training. It is thus important to see how well the model performs when only a fraction of the dense trajectories is available.

Figure~\ref{fig:pretrain-scalability} shows that even with limited dense data, the proposed model remains robust. Notably, using only 20\% of the dense pre-training data still leads to a significant performance gain over not using pre-training at all (0\%). This finding demonstrates the viability of our approach when large-scale dense data are hard to obtain.

\begin{figure*}
    \centering
    \pgfplotstableread[row sep=\\,col sep=&]{
scale & 1 & 2 & 4 \\
0 & 83.720 & 77.425 & 72.471 \\
20 & 87.906 & 83.382 & 78.406 \\
40 & 88.388 & 83.513 & 78.991 \\
60 & 88.932 & 83.664 & 79.212 \\
80 & 88.994 & 84.077 & 79.469 \\
100 & 89.064 & 84.361 & 80.839 \\
}\ptScaleStrPrec

\pgfplotstableread[row sep=\\,col sep=&]{
scale & 1 & 2 & 4 \\
0 & 194.30 & 272.86 & 479.49 \\
20 & 151.33 & 218.02 & 358.03 \\
40 & 142.61 & 204.67 & 348.69 \\
60 & 134.53 & 197.44 & 331.36 \\
80 & 133.95 & 196.18 & 317.08 \\
100 & 133.46 & 192.66 & 305.08 \\
}\ptScaleStrMAE

\pgfplotstableread[row sep=\\,col sep=&]{
scale & 1 & 2 & 4 \\
0   & 71.795 & 50.041 & 33.882 \\
20  & 79.580 & 73.760 & 70.530 \\
40  & 81.790 & 76.600 & 71.240 \\
60  & 82.120 & 77.150 & 71.350 \\
80  & 82.660 & 78.040 & 72.060 \\
100 & 83.052 & 79.236 & 72.560 \\
}\ptScaleTpAcc

\pgfplotstableread[row sep=\\,col sep=&]{
scale & 1 & 2 & 4 \\
0   & 376.05 & 540.89 & 857.43 \\
20  & 309.72 & 345.88 & 469.83 \\
40  & 289.48 & 316.16 & 399.52 \\
60  & 280.84 & 313.19 & 362.44 \\
80  & 265.86 & 307.64 & 358.17 \\
100 & 257.19 & 297.07 & 352.54 \\
}\ptScaleTpMAE

\tikzset{every plot/.style={line width=1.2pt}}

\subfigure[Travel Time Estimation] {
    \resizebox{0.315\linewidth}{!}{
        \begin{tikzpicture}
        \begin{axis}[
            width=0.35\linewidth,
            ylabel={MAE (minutes)},
            ymin=0.555, ymax=0.68,
            xtick distance={20},
            ytick distance={0.025},
            ymajorgrids=true,
            grid style=dashed,
            nodes near coords,
            nodes near coords align={vertical},
            nodes near coords style={font=\small},
            every node near coord/.append style={anchor=south},
        ]
        \addplot[color=pttte,mark=square*]
            coordinates {
                (0,0.666)
                (20,0.575)
                (40,0.563)
                (60,0.564)
                (80,0.560)
                (100,0.558)
            };
        \end{axis}
        \end{tikzpicture}
        
        \begin{tikzpicture}
        \begin{axis}[
            width=0.35\linewidth,
            ylabel={MAPE (\%)},
            ymin=8.7, ymax=10.65,
            xtick distance={20},
            ytick distance={0.4},
            ymajorgrids=true,
            grid style=dashed,
            nodes near coords,
            nodes near coords align={vertical},
            nodes near coords style={font=\small},
            every node near coord/.append style={anchor=south},
        ]
        \addplot[color=pttte,mark=square*]
            coordinates {
                (0,10.501)
                (20,9.297)
                (40,9.230)
                (60,9.105)
                (80,8.989)
                (100,8.745)
            };
        \end{axis}
        \end{tikzpicture}
    }
    \label{fig:pretrain-scalability-travel-time-estimation}
}
\subfigure[Trajectory Recovery] {
    \resizebox{0.315\linewidth}{!}{
        \begin{tikzpicture}
        \begin{axis}[
            width=0.35\linewidth,
            ylabel={Precision (\%)},
            ymin=72, ymax=90.5,
            xtick distance={20},
            ytick distance={4},
            ymajorgrids=true,
            grid style=dashed,
            nodes near coords,
            nodes near coords align={vertical},
            nodes near coords style={font=\small},
            every node near coord/.append style={anchor=south},
        ]
        \addplot[color=pt1min,mark=square*] table[x=scale,y=1]{\ptScaleStrPrec};
        \addplot[color=pt2min,mark=triangle*] table[x=scale,y=2]{\ptScaleStrPrec};
        \addplot[color=pt4min,mark=otimes*] table[x=scale,y=4]{\ptScaleStrPrec};
        \end{axis}
        \end{tikzpicture}
        
        \begin{tikzpicture}
        \begin{axis}[
            width=0.35\linewidth,
            ylabel={MAE (Coor, meters)},
            ymin=125, ymax=510,
            xtick distance={20},
            ytick distance={80},
            ymajorgrids=true,
            grid style=dashed,
            nodes near coords,
            nodes near coords align={vertical},
            nodes near coords style={font=\small},
            every node near coord/.append style={anchor=south},
        ]
        \addplot[color=pt1min,mark=square*] table[x=scale,y=1]{\ptScaleStrMAE};
        \addplot[color=pt2min,mark=triangle*] table[x=scale,y=2]{\ptScaleStrMAE};
        \addplot[color=pt4min,mark=otimes*] table[x=scale,y=4]{\ptScaleStrMAE};
        \end{axis}
        \end{tikzpicture}
    }
    \label{fig:pretrain-scalability-sparse-trajectory-recovery}
}
\subfigure[Trajectory Prediction] {
    \resizebox{0.315\linewidth}{!}{
        \begin{tikzpicture}
        \begin{axis}[
            width=0.35\linewidth,
            ylabel={Accuracy (\%)},
            ymin=32, ymax=87,
            xtick distance={20},
            ytick distance={10},
            ymajorgrids=true,
            grid style=dashed,
            nodes near coords,
            nodes near coords align={vertical},
            nodes near coords style={font=\small},
            every node near coord/.append style={anchor=south},
        ]
        \addplot[color=pt1min,mark=square*] table[x=scale,y=1]{\ptScaleTpAcc};
        \addplot[color=pt2min,mark=triangle*] table[x=scale,y=2]{\ptScaleTpAcc};
        \addplot[color=pt4min,mark=otimes*] table[x=scale,y=4]{\ptScaleTpAcc};
        \end{axis}
        \end{tikzpicture}
        \begin{tikzpicture}
        \begin{axis}[
            width=0.35\linewidth,
            ylabel={MAE (Coor, meters)},
            ymin=245, ymax=910,
            xtick distance={20},
            ytick distance={120},
            ymajorgrids=true,
            grid style=dashed,
            nodes near coords,
            nodes near coords align={vertical},
            nodes near coords style={font=\small},
            every node near coord/.append style={anchor=south},
        ]
        \addplot[color=pt4min,mark=otimes*] table[x=scale,y=4]{\ptScaleTpMAE};
        \addplot[color=pt2min,mark=triangle*] table[x=scale,y=2]{\ptScaleTpMAE};
        \addplot[color=pt1min,mark=square*] table[x=scale,y=1]{\ptScaleTpMAE};
        \legend{$\mu=1$ minute,$\mu=2$ minutes,$\mu=4$ minutes}
        \end{axis}
        \end{tikzpicture}
    }
    \label{fig:pretrain-scalability-trajectory-prediction}
}
    \caption{\rev{Scalability with regard to the size (\%) of pre-train sets in Chengdu.}}
    \label{fig:pretrain-scalability}
\end{figure*}

\subsubsection{Scalability of Fine-tuning with Limited Task-specific Data}
Comparison in Tables~\ref{tab:travel-time-estimation-accuracy} and~\ref{tab:trajectory-prediction-accuracy} show that fine-tuning the model further improves performance, given that the input format used in each task may differ from that in pre-training. This raises the question of whether extensive fine-tuning data is needed to achieve optimal results.

Figure~\ref{fig:finetune-scalability} illustrates that the model can approach its peak performance with just 20\% of the task-specific data. Even a small fine-tuning set significantly boosts accuracy compared to no fine-tuning (0\%). These results highlight the model’s practicality in real-world settings where obtaining large labeled datasets for fine-tuning can be difficult.

\begin{figure*}
    \centering
    \pgfplotstableread[row sep=\\,col sep=&]{
scale & 1 & 2 & 4 \\
0   & 87.658 & 83.605 & 79.102 \\
20  & 88.355 & 84.010 & 79.315 \\
40  & 88.673 & 84.276 & 79.960 \\
60  & 88.998 & 84.317 & 80.177 \\
80  & 89.050 & 84.338 & 80.627 \\
100 & 89.064 & 84.361 & 80.839 \\
}\ftScaleStrPrec

\pgfplotstableread[row sep=\\,col sep=&]{
scale & 1 & 2 & 4 \\
0   & 137.82 & 218.33 & 350.51 \\
20  & 135.57 & 202.16 & 313.00 \\
40  & 135.07 & 198.61 & 310.12 \\
60  & 134.15 & 193.05 & 306.82 \\
80  & 133.90 & 194.22 & 305.30 \\
100 & 133.46 & 192.66 & 305.08 \\
}\ftScaleStrMAE

\pgfplotstableread[row sep=\\,col sep=&]{
scale & 1 & 2 & 4 \\
0   & 68.912 & 63.273 & 55.011 \\
20  & 80.015 & 73.881 & 68.209 \\
40  & 81.092 & 77.024 & 70.776 \\
60  & 82.020 & 78.426 & 71.754 \\
80  & 82.727 & 79.105 & 72.279 \\
100 & 83.052 & 79.236 & 72.560 \\
}\ftScaleTpAcc

\pgfplotstableread[row sep=\\,col sep=&]{
scale & 1 & 2 & 4 \\
0   & 324.34 & 390.73 & 463.14 \\
20  & 288.12 & 342.17 & 403.12 \\
40  & 280.09 & 327.08 & 389.54 \\
60  & 270.77 & 308.99 & 371.83 \\
80  & 263.21 & 300.85 & 358.19 \\
100 & 257.19 & 297.07 & 352.54 \\
}\ftScaleTpMAE

\tikzset{every plot/.style={line width=1.2pt}}

\subfigure[Travel Time Estimation] {
    \resizebox{0.315\linewidth}{!}{
        \begin{tikzpicture}
        \begin{axis}[
            width=0.35\linewidth,
            ylabel={MAE (minutes)},
            ymin=0.55, ymax=0.75,
            xtick distance={20},
            ytick distance={0.05},
            ymajorgrids=true,
            grid style=dashed,
            nodes near coords,
            nodes near coords align={vertical},
            nodes near coords style={font=\small},
            every node near coord/.append style={anchor=south},
        ]
        \addplot[color=pttte,mark=square*]
            coordinates {
                (0,0.725)
                (20,0.620)
                (40,0.595)
                (60,0.582)
                (80,0.569)
                (100,0.558)
            };
        \end{axis}
        \end{tikzpicture}
        \begin{tikzpicture}
        \begin{axis}[
            width=0.35\linewidth,
            ylabel={MAPE (\%)},
            ymin=8.5, ymax=11.5,
            xtick distance={20},
            ytick distance={0.6},
            ymajorgrids=true,
            grid style=dashed,
            nodes near coords,
            nodes near coords align={vertical},
            nodes near coords style={font=\small},
            every node near coord/.append style={anchor=south},
        ]
        \addplot[color=pttte,mark=square*]
            coordinates {
                (0,11.148)
                (20,9.724)
                (40,9.424)
                (60,9.317)
                (80,9.104)
                (100,8.745)
            };
        \end{axis}
        \end{tikzpicture}
    }
    \label{fig:finetune-scalability-travel-time-estimation}
}
\subfigure[Trajectory Recovery] {
    \resizebox{0.315\linewidth}{!}{
        \begin{tikzpicture}
        \begin{axis}[
            width=0.35\linewidth,
            ylabel={Precision (\%)},
            ymin=78.5, ymax=90,
            xtick distance={20},
            ytick distance={2.5},
            ymajorgrids=true,
            grid style=dashed,
            nodes near coords,
            nodes near coords align={vertical},
            nodes near coords style={font=\small},
            every node near coord/.append style={anchor=south},
        ]
        \addplot[color=pt1min,mark=square*] table[x=scale,y=1]{\ftScaleStrPrec};
        \addplot[color=pt2min,mark=triangle*] table[x=scale,y=2]{\ftScaleStrPrec};
        \addplot[color=pt4min,mark=otimes*] table[x=scale,y=4]{\ftScaleStrPrec};
        \end{axis}
        \end{tikzpicture}
        \begin{tikzpicture}
        \begin{axis}[
            width=0.35\linewidth,
            ylabel={MAE (Coor, meters)},
            ymin=125, ymax=370,
            xtick distance={20},
            ytick distance={50},
            ymajorgrids=true,
            grid style=dashed,
            nodes near coords,
            nodes near coords align={vertical},
            nodes near coords style={font=\small},
            every node near coord/.append style={anchor=south},
        ]
        \addplot[color=pt1min,mark=square*] table[x=scale,y=1]{\ftScaleStrMAE};
        \addplot[color=pt2min,mark=triangle*] table[x=scale,y=2]{\ftScaleStrMAE};
        \addplot[color=pt4min,mark=otimes*] table[x=scale,y=4]{\ftScaleStrMAE};
        \end{axis}
        \end{tikzpicture}
    }
    \label{fig:finetune-scalability-sparse-trajectory-recovery}
}
\subfigure[Trajectory Prediction] {
    \resizebox{0.315\linewidth}{!}{
        \begin{tikzpicture}
        \begin{axis}[
            width=0.35\linewidth,
            ylabel={Accuracy (\%)},
            ymin=54, ymax=86,
            xtick distance={20},
            ytick distance={7},
            ymajorgrids=true,
            grid style=dashed,
            nodes near coords,
            nodes near coords align={vertical},
            nodes near coords style={font=\small},
            every node near coord/.append style={anchor=south},
            legend pos=south east
        ]
        \addplot[color=pt1min,mark=square*] table[x=scale,y=1]{\ftScaleTpAcc};
        \addplot[color=pt2min,mark=triangle*] table[x=scale,y=2]{\ftScaleTpAcc};
        \addplot[color=pt4min,mark=otimes*] table[x=scale,y=4]{\ftScaleTpAcc};
        \legend{$\mu=1$ minute,$\mu=2$ minutes,$\mu=4$ minutes}
        \end{axis}
        \end{tikzpicture}
        \begin{tikzpicture}
        \begin{axis}[
            width=0.35\linewidth,
            ylabel={MAE (Coor, meters)},
            ymin=250, ymax=480,
            xtick distance={20},
            ytick distance={60},
            ymajorgrids=true,
            grid style=dashed,
            nodes near coords,
            nodes near coords align={vertical},
            nodes near coords style={font=\small},
            every node near coord/.append style={anchor=south},
        ]
        \addplot[color=pt1min,mark=square*] table[x=scale,y=1]{\ftScaleTpMAE};
        \addplot[color=pt2min,mark=triangle*] table[x=scale,y=2]{\ftScaleTpMAE};
        \addplot[color=pt4min,mark=otimes*] table[x=scale,y=4]{\ftScaleTpMAE};
        \end{axis}
        \end{tikzpicture}
    }
    \label{fig:finetune-scalability-trajectory-prediction}
}
    \caption{\rev{Scalability with regard to the size (\%) of fine-tune sets in Chengdu.}}
    \label{fig:finetune-scalability}
\end{figure*}

\subsubsection{Effectiveness of Hyper-parameters} \label{sec:hyper-parameters}

\begin{figure*}
    \centering
    \input{plots/hyper-parameters}
    \caption{\rev{Effectiveness of hyper-parameters on the trajectory recovery task in Chengdu.}}
    \label{fig:hyper-parameters}
\end{figure*}

We study how the five main hyper-parameters in Table~\ref{tab:optimal-parameter} affect our model’s performance. The results are shown in Figure~\ref{fig:hyper-parameters}. Key observations include:
\begin{itemize}[leftmargin=*]
    \item \textbf{Embedding dimensionality $d$.} As seen in Figure~\ref{fig:hyper-parameter-d}, increasing $d$ generally improves performance. Beyond $d=128$, however, gains become marginal while computational cost continues to rise. Hence, we choose $d=128$ as a good balance.
    \item \textbf{Number of attention heads $N_h$.} From Figure~\ref{fig:hyper-parameter-N_h}, both Precision and MAE are optimal at $N_h=8$.
    \item \textbf{Distance threshold $\delta$.} This threshold controls how many neighboring road segments are considered. Figure~\ref{fig:hyper-parameter-delta} shows that it mainly affects Precision for road segments, with best results at $\delta=100$. A smaller $\delta$ can miss possible segments, while a larger one may add noise.
    \item \rev{\textbf{Feature removal probability $\varphi$.} This parameter determines how often features are dropped during pre-training. As shown in Figure~\ref{fig:hyper-parameter-phi}, a 20\% setting yields the best balance between reconstruction difficulty and performance.}
    \item \rev{\textbf{Batch size $|\mathbb T|$.} Figure~\ref{fig:hyper-parameter-B} indicates that a larger batch can improve performance of contrastive learning, but returns diminish past $|\mathbb T|=256$. We therefore set $|\mathbb T|=256$.}
\end{itemize}

\subsubsection{Effectiveness of Modules}
We run an ablation study to determine how each feature and module affects performance, comparing the full model against these variants:
\begin{itemize}[leftmargin=*]
    \item \textbf{w/o neigh.}: Omits the road segment neighbor set $\Omega(l_i)$ in Equation~\ref{eq:tuple-of-feature-domains}.
    \item \textbf{w/o coor.}: Omits the coordinate feature $l_i$ in Equation~\ref{eq:tuple-of-feature-domains}.
    \item \textbf{w/o time}: Omits the time feature $t_i$ in Equation~\ref{eq:tuple-of-feature-domains}.
    \item \textbf{w/o shuffle}: Disables the shuffling of generated spans during pre-training.
    \item \textbf{Flat encoder}: Computes $\boldsymbol h_i$ by directly mean-pooling $\boldsymbol Z_i$, instead of applying self-attention.
    \item \textbf{FC num. enc.}: Replaces the continuous feature encoder $\varPhi$ with a fully connected network.
\end{itemize}

\begin{table}
    \centering
    \caption{Effectiveness of features and modules on the trajectory recovery task in Chengdu.}
    \resizebox{1.0\linewidth}{!}{
\begin{threeparttable}
\begin{tabular}{c|cc}
\toprule
Sampling Interval $\mu$ & \multicolumn{2}{c}{1 minute / 2 minutes / 4 minutes} \\
\midrule
Variations & Precision (\%) $\uparrow$ & MAE (Coor, meters) $\downarrow$ \\
\midrule
w/o neigh. 
  & \rev{76.847 / 74.512 / 71.923} 
  & \rev{128.52 / 193.57 / \textbf{300.62}} \\

w/o coor. 
  & \rev{78.882 / 77.298 / 72.467} 
  & \rev{161.89 / 283.62 / 445.14} \\

w/o time 
  & \rev{86.779 / 83.265 / 77.092} 
  & \rev{141.23 / 203.08 / 319.33} \\

w/o shuffle 
  & \rev{84.413 / 77.455 / 73.615} 
  & \rev{159.06 / 275.99 / 433.61} \\

Flat encoder 
  & \rev{85.573 / 79.902 / 74.146} 
  & \rev{141.88 / 217.02 / 323.98} \\

FC num. enc. 
  & \rev{86.039 / 81.451 / 76.929} 
  & \rev{145.52 / 218.12 / 327.48} \\
\textbf{UVTM} & \rev{\textbf{89.064} / \textbf{84.361} / \textbf{80.839}} & \rev{\textbf{133.46} / \textbf{192.66} / 305.08} \\
\bottomrule
\end{tabular}
\begin{tablenotes}\footnotesize
\item[]{
    \textbf{Bold} denotes the best result. $\uparrow$ means higher is better, and $\downarrow$ means lower is better.
}
\end{tablenotes}
\end{threeparttable}
}
    \label{tab:modules}
\end{table}

The results are computed on Chengdu's test set for the TR task, as shown in Table~\ref{tab:modules}. We draw the following observations:
\begin{itemize}[leftmargin=*]
    \item Road-segment neighbors, coordinates, and time features all contribute to improved performance. Removing any one of these features leads to a clear drop.
    \item Not shuffling the generated spans in pre-training hurts the model’s ability to capture bidirectional correlations in trajectories.
    \item The hierarchical attention in the trajectory encoder and the learnable feature encoder both increase the model’s accuracy.
\end{itemize}

\section{Conclusion} \label{sec:conclusion}
\rev{
We present a universal vehicle trajectory model (UVTM) that can be trained once and then adapted to a variety of trajectory-related tasks. With UVTM, there is no longer a need to maintain and update multiple task-specific models, which significantly reduces computational and storage overhead.
}

\rev{
To handle scenarios where trajectory features are incomplete---including tasks involving partially available features and sparse trajectories---UVTM introduces two specialized mechanisms. First, it divides trajectory features into three domains: spatial, temporal, and road. Each domain can be masked and generated independently, allowing flexible handling of partial features. Second, UVTM is pre-trained to reconstruct dense, feature-complete trajectories from sparse, feature-incomplete ones, preserving its effectiveness when facing incomplete trajectories.
}

\rev{
We conduct extensive experiments on three real-world vehicle GPS trajectory datasets and four representative tasks. Comparisons with state-of-the-art task-specific and universal trajectory models demonstrate that UVTM achieves superior performance across all tasks.
}

\ifCLASSOPTIONcompsoc
  \section*{Acknowledgments}
\else
  \section*{Acknowledgment}
\fi

This work was supported by the National Natural Science Foundation of China (No. 62272033).

\ifCLASSOPTIONcaptionsoff
  \newpage
\fi

\bibliographystyle{IEEEtran}
\bibliography{reference}

\begin{IEEEbiography}[{\includegraphics[width=1in,clip,keepaspectratio]{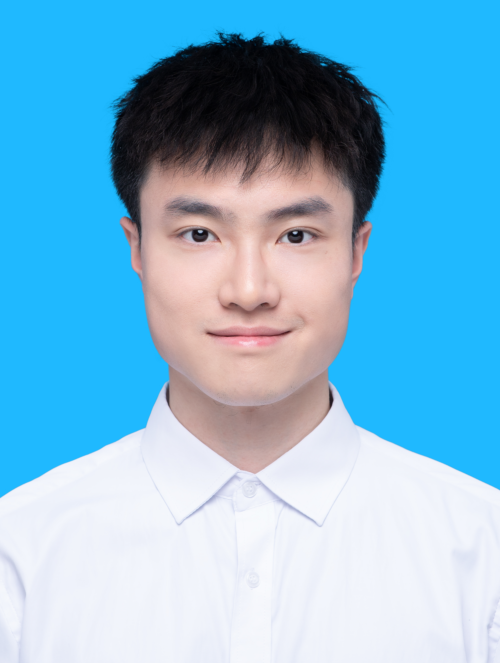}}]{Yan Lin} received the Ph.D. degree in computer science from Beijing Jiaotong University, Beijing, China, in 2024.

He is a Postdoc at the Department of Computer Science, Aalborg University. His research interests include spatiotemporal data mining and representation learning.
\end{IEEEbiography}

\begin{IEEEbiography}[{\includegraphics[width=1in,clip,keepaspectratio]{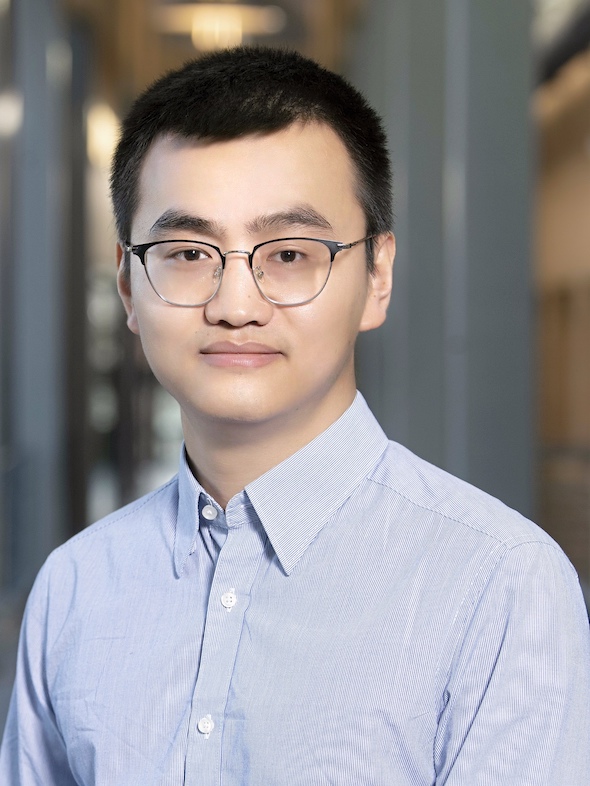}}]{Jilin Hu} received the Ph.D. degree in computer science from Aalborg University, Aalborg, Denmark in 2019.

He is a Professor at the School of Data Science and Engineering, East China Normal University. 
His research interests include spatiotemporal data analytics and transportation data mining.
\end{IEEEbiography}

\begin{IEEEbiography}[{\includegraphics[width=1in,clip,keepaspectratio]{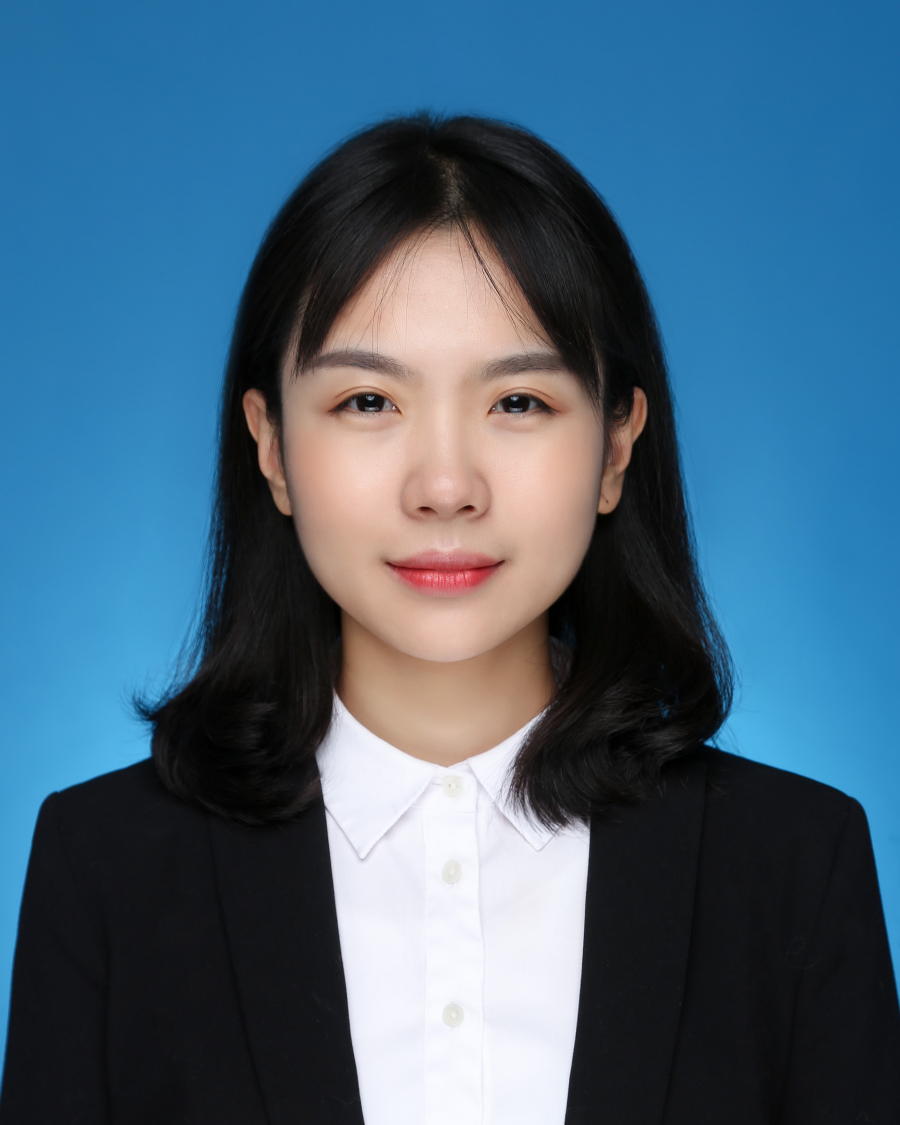}}]{Shengnan Guo} received the Ph.D. degree in computer science from Beijing Jiaotong University, Beijing, China, in 2021.

She is an associate professor at the School of Computer Science and Technology, Beijing Jiaotong University. Her research interests focus on spatial-temporal data mining and intelligent transportation systems.
\end{IEEEbiography}

\begin{IEEEbiography}[{\includegraphics[width=1in,clip,keepaspectratio]{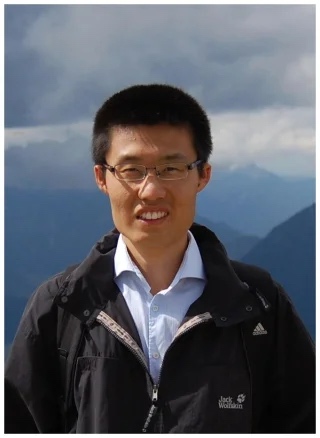}}]{Bin Yang} received the Ph.D. degree in computer science from Fudan University, Shanghai, China in 2010.

He is a Professor at the School of Data Science and Engineering, East China Normal University. 
His research interests include spatiotemporal data analytics, machine learning, and data management.
\end{IEEEbiography}

\begin{IEEEbiography}[{\includegraphics[width=1in,clip,keepaspectratio]{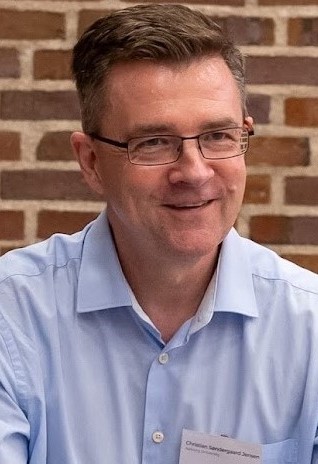}}]{Christian S. Jensen} received the Ph.D. degree from Aalborg University in 1991 after 2 1/2 years of study at University of Maryland, and he received the Dr.Techn. degree from Aalborg University in 2000.

He is a Professor at the Department of Computer Science, Aalborg University. 
His research concerns primarily temporal and spatiotemporal data management and analytics, including indexing and query processing, data mining, and machine learning.
\end{IEEEbiography}

\begin{IEEEbiography}[{\includegraphics[width=1in,clip,keepaspectratio]{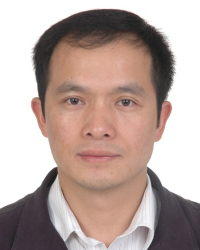}}]{Youfang Lin} received the Ph.D. degree in signal and information processing  from Beijing Jiaotong University, Beijing, China, in 2003.

He is a Professor with the School of Computer Science and Technology, Beijing Jiaotong University. His main fields of expertise and current research interests include big data technology, intelligent systems, complex networks, and traffic data mining.
\end{IEEEbiography}

\begin{IEEEbiography}[{\includegraphics[width=1in,clip,keepaspectratio]{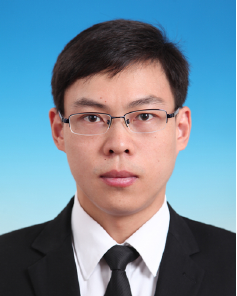}}]{Huaiyu Wan} received the Ph.D. degree in computer science and technology from Beijing Jiaotong University, Beijing, China, in 2012.

He is a Professor with the School of Computer Science and Technology, Beijing Jiaotong University. His current research interests focus on spatiotemporal data mining, social network mining, information extraction, and knowledge graph.
\end{IEEEbiography}


\end{document}